\documentclass[runningheads]{comsis2}

\def\journalissue{}
\def\paperidnum{}
\usepackage[square,numbers]{natbib} % has a nice set of citation styles and commands
\usepackage{geometry}
\usepackage[colorlinks=true,bookmarks=true,citecolor=blue,urlcolor=blue]{hyperref} %pdflatex
\usepackage{microtype}
\usepackage{graphicx}
\usepackage{subfig}
\usepackage{booktabs} % for professional tables
\usepackage{paralist}
\usepackage{soul}
\usepackage{algorithmic}
\usepackage[linesnumbered,lined,boxed,commentsnumbered,ruled,vlined]{algorithm2e}

\usepackage{float}
\usepackage[british]{babel}
\usepackage{amsmath}
\usepackage{amssymb}
\usepackage{mathtools}
\usepackage{stackrel}
\usepackage{apxproof}
\usepackage{url}            % simple URL typesetting
\usepackage{booktabs}       % professional-quality tables
\usepackage{amsfonts}       % blackboard math symbols
\usepackage{nicefrac}       % compact symbols for 1/2, etc.
\usepackage{xcolor}         % colors
\usepackage{mathptmx}
\usepackage{dsfont}
\usepackage[title]{appendix}
\usepackage{multirow}
\usepackage{wrapfig}
\DeclareMathAlphabet{\mathcal}{OMS}{cmsy}{m}{n}
\usepackage{tabularx}
\usepackage{bm}
\usepackage{tikz} 
\tikzset{
  agent/.style={draw, circle, minimum size=2mm,
  font=\footnotesize},
  x = 1.5cm, y=1.5cm,
  every loop/.style={},
  block/.style={draw, rectangle, minimum height=3em, minimum width=6em},
}
\usetikzlibrary{decorations.pathreplacing}
\usetikzlibrary{positioning}
\usetikzlibrary{shapes,arrows,matrix,scopes,shadows,chains,automata,positioning,fit, arrows.meta,calc}

\usepackage[capitalize,noabbrev]{cleveref}
\usepackage{libertine}
\usepackage{notation}
\crefname{subsection}{subsection}{subsections}
\crefname{lemma}{lemma}{lemma}
\crefname{property}{property}{property}
\crefname{table}{table}{table}
\crefname{assumption}{assumption}{assumption}
\crefname{table}{table}{tables}
\Crefname{table}{Table}{Tables}
\crefname{figure}{Fig.}{Fig.}
\Crefname{figure}{Fig.}{Fig.}

\newtheorem{assumption}[theorem]{Assumption}
\title{On learning history based policies for controlling Markov decision processes}
\titlerunning{On learning history based policies for controlling Markov decision processes}

\author{Gandharv Patil\inst{1} \and Aditya Mahajan\inst{1} \and Doina Precup\inst{1}}

\institute{McGill University,  Mila\\
  Montreal QC - Canada\\
  \email{gandharv.patil@mail.mcgill.ca,  aditya.mahajan@mcgill.ca,  dprecup@cs.mcgill.ca}}
\begin{document}

\maketitle

\begin{abstract}
Reinforcement learning (RL) folklore suggests that history-based function approximation methods, such as recurrent neural nets or history-based state abstraction, perform better than their memory-less counterparts, due to the fact that function approximation in Markov decision processes (MDP) can be viewed as inducing a Partially observable MDP. However, there has been little formal analysis of such history-based algorithms, as most existing frameworks focus exclusively on memory-less features. In this paper, we introduce a theoretical framework for studying the behaviour of RL algorithms that learn to control an MDP using history-based feature abstraction mappings. Furthermore, we use this framework to design a practical RL algorithm and we numerically evaluate its effectiveness on a set of continuous control tasks.

\end{abstract}

\section{Introduction}\label{sec:intro}
State abstraction and function approximation are vital components used by reinforcement learning (RL) algorithms to efficiently solve complex control problems  when exact computations are intractable due to large state and action spaces. Over the past few decades, state abstraction in RL has evolved from the use of pre-determined and problem-specific features~\cite{CritesB95, TsitsiklisR96,ndp,Sutton+Barto:1998,SinghLKW02,activesensing, ProperT06} to the use of adaptive basis functions learnt by solving an isolated regression problem~\citep{kbrl,autobasis-MenacheMS05,6-keller,Petrik07}, and more recently to the use of neural network-based Deep-RL algorithms that embed state abstraction in successive layers of a neural network~\citep{Barto2004SynthesisON,BellemareDDTCRS19}.
 
Feature abstraction results in information loss, and the resulting state features might not satisfy the controlled Markov property, even if this property is satisfied by the corresponding state ~\citep{Sutton+Barto:2018}. One approach to counteract the loss of the Markov property is to generate the features using the history of state-action pairs, and empirical evidence suggests that using such history-based features are beneficial in practice~\citep{openai2019learning}. However, a theoretical characterisation of history-based Deep-RL algorithms for fully observed Markov Decision Processes (MDPs) is largely absent form the literature. 

% However, besides the general Partially Observable(PO)-MDP assumptions, the theoretical characterisation of history-based RL algorithms for controlling MDPs is absent from the literature.
In this paper, we bridge this gap between theory and practise by providing a theoretical analysis of history-based RL agents acting in a MDP.
% This paper introduces a theoretical framework to analyse the use of history-based feature abstraction in RL agents acting in Markov Decision Processes (MDPs). 
Our approach adapts the notion of approximate information state (AIS) for POMDPs proposed in \cite{ais-1,ais-2} to feature abstraction in MDPs, and we develop a theoretically grounded policy search algorithm for history-based feature abstractions and policies. 
% and evaluate it on continuous control tasks. 

The rest of the paper is organised as follows: In \cref{sec:background},
following a brief review of feature-based abstraction, we motivate the need for using history-based feature abstractions. In \cref{sec:main}, we present a formal model for the co-design of the feature abstraction and control policy, derive a dynamic program using the AIS. We also derive bounds on the quality of approximate solutions to this dynamic program. 
In \Cref{sec:algorithm} we build on these approximation bounds to develop an RL algorithm for learning a history-based state representation and control policy. In \cref{sec:experiments}, we present an empirical evaluation of our proposed algorithm on continuous control tasks. Finally, we discuss related work in \cref{sec:litreview} and conclude with future research directions in \cref{sec:conclusion}.

\section{Background and Motivation}\label{sec:background}

    % \subsection{Function approximation in MDPs}
    
        Consider an MDP $ \mdp = \langle \statespace, \actionspace, \transition, \cost, \discount \rangle$ where $\statespace$ denotes the state space, $\actionspace$ denotes the action space, $\transition$ denotes the controlled transition matrix, $\cost \colon \statespace \times \actionspace \to \real$ denotes the per-step reward, and $\discount \in (0,1)$ denotes the discount factor.
                
        The performance of a randomised (and possibly history-dependent) policy $\policy$
        starting from a start state $\sts_{0}$ is measured by the value function, defined as:
        \begin{equation}
          \valuefunction^{\policy}(\sts_0) = \expecun{}^{\policy}\bigg[\sum_{\timestep=1}^{\infty}\discount^{\timestep-1}\cost(\State_{\timestep},\Action_{\timestep}) \bigg| \State_{0} = \sts_{0}\bigg].
        \end{equation}
        A policy maximising $\valuefunction^{\policy}(\sts_0)$ over all (randomised and possibly history dependent) policies is called the optimal policy with respect to initial state $s_0$ and is denoted by $\policy^{\star}$.
        %\footnote{This notion can easily be extended to start state distributions.}.
                
        In many applications, $\statespace$ and $\actionspace$ are combinatorially large or uncountable, which makes it intractable to compute the optimal policy. 
        Most practical RL algorithms overcome this hurdle by using function approximation where the state is mapped to a feature space $\featurespace$ using a state abstraction function $\basis:\statespace \to \featurespace$. In Deep-RL algorithms, the last layer of the network is often viewed as a feature vector. These feature vectors are then used as an approximate state for approximating the value function $\hat\valuefunction: \featurespace \to \real$ and/or computing an approximately optimal policy $\mu:\featurespace \to \Delta(\actionspace)$~\citep{Sutton+Barto:1998} (where $\Delta(\actionspace)$ denotes the set of probability distribution over actions). Therefore, the mapping from state to distribution of actions is given by the ``flattened'' policy $\tilde{\mu} = \mu \circ \basis$ ~\ie, $\tilde{\mu} = \mu(\phi(\cdot))$.
        
        % The features $\Feature_\timestep$ are then used as an approximate state for computing the value function $\hat\valuefunction: \featurespace \to \real$ and/or the policy $\mu:\featurespace \to \Delta(\actionspace)$~\citep{Sutton+Barto:1998} (where $\Delta(\actionspace)$ denotes the set of probability distribution over actions). We denote by $\mu \circ \phi $ as the ``flattened" policy that maps states to action distributions, by composing the feature abstraction and the policy acting on it. In Deep-RL algorithms, the output of the last layer of the network is often viewed as the feature vector $\Feature_{\timestep}$.
        
        % A well known yet often overlooked fact about function approximation is that the features used as a proxy state may not satisfy the controlled Markov property \ie, in general,
        A well known fact about function approximation is that the features that are used as an approximate state may not satisfy the controlled Markov property \ie, in general, 
        \[
            \prob(\Feature_{\timestep+1} \mid \Feature_{1:\timestep}, \Action_{1:\timestep}) \neq
            \prob(\Feature_{\timestep+1} \mid \Feature_\timestep, \Action_\timestep).
        \]

        \begin{figure}[!h]
          \centering
          \begin{minipage}{\linewidth}
          \subfloat[$P(0)$]{
            \begin{tikzpicture}[thick,scale=0.9]
              \node [agent] at (0, 0) (0) {$0$};
              \node [agent] at (1, 0) (1) {$1$}; 
              \node [agent] at (0, 1) (3) {$3$};
              \node [agent] at (1, 1) (2) {$2$};
              \path[->]
                    (0) edge node[below] {$0.5$} (1)
                    (1) edge node[right] {$0.5$} (2)
                    (2) edge node[above] {$0.5$} (3)
                    (3) edge node[left]  {$0.5$} (0)
                    (0) edge[loop left] node {$0.5$} (0)
                    (1) edge[loop right] node {$0.5$} (1)
                    (2) edge[loop right] node {$0.5$} (2)
                    (3) edge[loop left] node {$0.5$} (3)
                    ;
            \end{tikzpicture}
            \label{fig:P(1)}}
           \hfill
          \subfloat[$P(1)$]{
            % \centering
            \begin{tikzpicture}[thick,scale=0.9]
              \node [agent] at (0, 0) (0) {$0$};
              \node [agent] at (1, 0) (1) {$1$}; 
              \node [agent] at (0, 1) (3) {$3$};
              \node [agent] at (1, 1) (2) {$2$};
              \path[<-]
                    (0) edge node[below] {$0.5$} (1)
                    (1) edge node[right] {$0.5$} (2)
                    (2) edge node[above] {$0.5$} (3)
                    (3) edge node[left]  {$0.5$} (0)
                    (0) edge[loop left] node {$0.5$} (0)
                    (1) edge[loop right] node {$0.5$} (1)
                    (2) edge[loop right] node {$0.5$} (2)
                    (3) edge[loop left] node {$0.5$} (3)
                    ;
            \end{tikzpicture}
            \label{fig:P(2)}}
            \hfill
            \subfloat[$P(2)$]{%
                  \begin{tikzpicture}[thick,scale=0.9]
                  \node [agent] at (0, 0) (0) {$0$};
                  \node [agent] at (1, 0) (1) {$1$}; 
                  \node [agent] at (0, 1) (3) {$3$};
                  \node [agent] at (1, 1) (2) {$2$};
                  \path[->]
                        (0) edge [bend right] node[below] {$0.5$} (1)
                        (1) edge [bend right] node[right] {$0.5$} (2)
                        (2) edge [bend right] node[above] {$0.5$} (3)
                        (3) edge [bend right] node[left]  {$0.5$} (0)
                        (0) edge [bend right] node[right] {} (3)
                        (3) edge [bend right] node[below] {} (2)
                        (2) edge [bend right] node[left]  {} (1)
                        (1) edge [bend right] node[above]  {} (0);
                \end{tikzpicture}
                \label{fig:P(3)}}
            \hfill
            \subfloat[$P_{\policy{}}$]{
            \begin{tikzpicture}[thick,scale=0.9]
              \node [agent] at (0, 0) (0) {$0$};
              \node [agent] at (1, 0) (1) {$1$}; 
              \node [agent] at (0, 1) (3) {$3$};
              \node [agent] at (1, 1) (2) {$2$};
              \path[->]
                    (0) edge node[below] {$0.5$} (1)
                    (1) edge[bend right] node[right] {$0.5$} (2)
                    (2) edge[bend right]  node[left] {$0.5$} (1)
                    (3) edge node[left]  {$0.5$} (0)
                    (3) edge node[above] {$0.5$} (2)
                    (0) edge[loop left] node {$0.5$} (0)
                    (1) edge[loop right] node {$0.5$} (1)
                    (2) edge[loop right] node {$0.5$} (2)
                    ;
            \end{tikzpicture}
            \label{fig:optimal}
            }
          \end{minipage}%
          \caption{The transition probability for an example MDP}
        \end{figure}
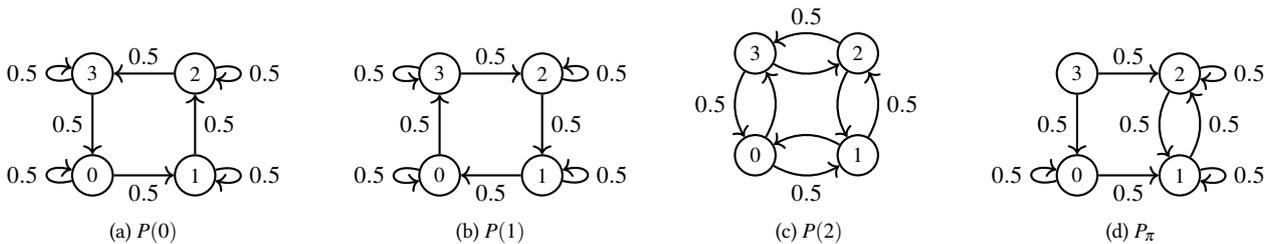

        To see the implications of this fact, consider the toy MDP depicted in~\cref{fig:P(1),fig:P(2),fig:P(3)}, with $ \statespace = \{0, 1, 2, 3\}$, $\actionspace =
        \{0, 1, 2\}$, $\{\transition_{\sts, \sts'}(\action)\}_{\action \in \actionspace}$, and $r(0) = r(1) = -1$, $r(2) = 1$, $r(3) = -K$, where $K$ is a large positive number.
       Given the reward structure the objective of the policy is to try to avoid state~$3$ and keep the agent at state~$2$ as much as possible. It is easy to see that the optimal policy is 
            \[
              \pi^\star(0) = 0, \quad
              \pi^\star(1) = 0, \quad
              \pi^\star(2) = 1,
              \text{ and}\quad
              \pi^\star(3) = 2.
            \]
            
        Note that if the initial state is not state~$3$ then an agent will never visit that state under the optimal policy. Furthermore, any policy which cannot prevent the agent from visiting state~$3$ will have a large negative value and, therefore, cannot be optimal.
        Now suppose the feature space $\featurespace = \{0, 1\}$. It is easy to see that for any Markovian feature-abstraction $\policyencoder{} \colon \statespace \to \featurespace$, no policy $\hat \pi \colon \featurespace \to \actionspace$ can prevent the agent from visiting state~$3$. Thus, the best policy when using Markovian feature abstraction will perform significantly worse than the optimal policy (which has direct access to the state).
    
        However, it is possible to construct a history-based feature-abstraction  $\policyencoder$ and a history-based control policy $\hat \pi$ that works with $\phi$ and is of the same quality as $\pi^\star$. For this, consider the following \emph{codebooks} (where the entries denoted by a dot do not matter):    
            
        % \begin{wrapfigure}{rt}{0.25\linewidth}
        %   \begin{minipage}{\linewidth}
        %   \centering
        %     \begin{tikzpicture}[thick,scale=0.9]
        %       \node [agent] at (0, 0) (0) {$0$};
        %       \node [agent] at (1, 0) (1) {$1$}; 
        %       \node [agent] at (0, 1) (3) {$3$};
        %       \node [agent] at (1, 1) (2) {$2$};
        %       \path[->]
        %             (0) edge node[below] {$0.5$} (1)
        %             (1) edge[bend right] node[right] {$0.5$} (2)
        %             (2) edge[bend right]  node[left] {$0.5$} (1)
        %             (3) edge node[left]  {$0.5$} (0)
        %             (3) edge node[above] {$0.5$} (2)
        %             (0) edge[loop left] node {$0.5$} (0)
        %             (1) edge[loop right] node {$0.5$} (1)
        %             (2) edge[loop right] node {$0.5$} (2)
        %             % (3) edge[loop left] node {$0.5$} (3)
        %             ;
        %     \end{tikzpicture}
        %   \caption{The transition probability of $\pi^\star$}
        %   \label{fig:optimal}
        %   \end{minipage}
        %   \vspace{-3mm}
        % \end{wrapfigure}
        
        The Markov chain induced by the optimal policy is shown in~\Cref{fig:optimal}. 
        Now define
        \begin{align*}
          F(1) &= \begin{bmatrix}
            0 & 1 & \cdot & \cdot \\
            \cdot & 0 & 1 & \cdot \\
            \cdot & \cdot & 0 & 1 \\
            1 & \cdot & \cdot & 0
          \end{bmatrix}, 
          &
          F(2) &= \begin{bmatrix}
            1 & \cdot & \cdot & 0 \\
            0 & 1 & \cdot & \cdot \\
            \cdot & 0 & 1 & \cdot \\
            \cdot & \cdot & 0 & 1
          \end{bmatrix} ,
          &
          F(3) &= \begin{bmatrix}
            \cdot & 0 & \cdot & 1 \\
            0 & \cdot & 1 & \cdot \\
            \cdot & 0 & \cdot & 1 \\
            0 & \cdot & 1 & \cdot
          \end{bmatrix},
          \\
          D(0) &= \begin{bmatrix}
            0 & 1 \\
            1 & 2 \\
            2 & 3 \\
            3 & 0 
          \end{bmatrix} ,
          &
          D(1) &= \begin{bmatrix}
            3 & 0 \\
            0 & 1 \\
            1 & 2 \\
            2 & 3 
          \end{bmatrix} ,
          &
          D(2) &= \begin{bmatrix}
            1 & 3 \\
            0 & 2 \\
            1 & 3 \\
            0 & 2 
          \end{bmatrix} .
        \end{align*}

        and consider the feature-abstraction policy
        \(
          Z_t =  F_{S_{t-1}, S_t}(A_{t-1})
        \)
        and a control policy $\mu$ which is a finite state machine with memory, where the memory $M_t$ that is updated as
        \(
          M_t = D_{M_{t-1}, Z_t}(A_{t-1})
        \)
        and the action $A_t$ is chosen as
        \(
          A_t = \pi(M_t),
        \)
        where $\pi \colon \statespace \to \Delta(\actionspace)$ is any pre-specified reference policy. It can be verified that if the system starts from a known initial state then $\mu \circ \basis = \pi$. 
         Thus, if we choose the reference policy $\pi=\pi^\star$, then
        the agent will never visit state~$3$ under $\mu \circ \basis$, in contrast to Markovian feature-abstraction policies where (as we argued before) state~$3$ is always visited.
        
        In the above example, we used the properties of the system dynamics and the reward function to design a history-based feature abstraction which outperforms memoryless feature abstractions. We are interested in developing such history-based feature abstractions using a learning framework when the system model is not known. We present such a construction in the next section.

    \section{Approximation bounds for history-based feature abstraction}\label{sec:main}
         
        The approximation results of our framework depend on the properties of metrics on probability spaces. We start with a brief overview of a general class of metrics known as Integral Probability Measures (IPMs)~\citep{ipm}; many of the commonly used metrics on probability spaces such as total variation (TV) distance, Wasserstein distance, and maximum-mean discrepency (MMD) are instances of IPMs. We then derive a general approximation bound that holds for general IPMs, and then specialize the bound to specific instances (TV, Wassserstein, and MMD).
        
        %In particular, we derive results for Total Variation (TV) distance, Wasserstein/Kantorovich-Rubinstein distance and Maximum-Mean Discrepancy (MMD) distance. All of these metrics are instances of Integral Probability Measures (IPMs)~\citep{ipm}---a class of metrics that have a dual characterisation. Viewing all the above metrics through the lens of IPMs allows us to derive a general approximation bound that holds for all the instances of IPMs, and then specialise the bound to specific instances \ie, TV, Wasserstein and MMD distance. Before defining our main model we will briefly describe the important properties of IPMs necessary for deriving our results.
        
        \subsection{Integral probability metrics (IPM)}
            \begin{definition}[~\citep{ipm}]\label{def:ipm}
                    Let $(\mathcal{E},\mathcal{G})$ be a measurable space and $\mathfrak{F}$ denote a class of uniformly bounded measurable functions on $(\mathcal{E},\mathcal{G})$. The integral probability metric between two probability distributions  $\nu_1, \nu_2 \in \mathcal{P}(\mathcal{E})$ with respect to the function class $\mathfrak{F}$ is defined as:
                    \begin{align}
                        \ipm(\nu_1, \nu_2) &= \sup_{f \in \mathfrak{F}}\bigg| \int_{\mathcal{E}}f d\nu_1  - \int_{\mathcal{E}}f d\nu_2\bigg|.\label{eq:def-ipm}
                    \end{align}
                    
                    For any function $f$ (not necessarily in $\mathfrak{F}$), the Minkowski functional $\rho_{\mathfrak{F}}$ associated with the metric $\ipm$ is defined as:
                    \begin{align}
                        \rho_{\mathfrak{F}}(f)&\define \inf\{\rho \in \real_{\geq 0}: \rho^{-1}f \in \mathfrak{F}\}.\label{eq:minkowski-functional}
                    \end{align}
                    
                    Eq.~\eqref{eq:minkowski-functional}, implies that that for any function $f$:
                    \begin{align}
                         \bigg|\int_{\mathcal{E}}fd\nu_1 - \int_{\mathcal{E}}fd\nu_2 \bigg|\leq \rho_{\mathfrak{F}}(f)\ipm(\nu_1,\nu_2). \label{eq:ipm-implication}
                     \end{align}\label{eq:ipm-function-diff}
                    
            \end{definition}
            \noindent In this paper, we use the following IPMs:
            \begin{compactitem}
                \item[1.]\label{def:tv-dist}{\bf Total Variation Distance}: If $\mathfrak{F}$ is chosen as $\mathfrak{F}^{\text{TV}} \define \{\frac{1}{2}\spn(f)$ = $\frac{1}{2}(\max(f)- \min(f))\}$, then $\ipm$ is the total variation distance, and its Minkowski functional is $\rho_{\mathfrak{F}^{\text{TV}}}(f) = \frac{1}{2}\spn(f)$. 
                \item[2.]\label{def:kr-dist}{\bf Wasserstein/Kantorovich-Rubinstein Distance}: If $\mathcal{E}$ is a metric space and $\mathfrak{F}$ is chosen as $\mathfrak{F}^{W} \define \{f: L_f \leq 1 \}$ (where $L_f$ denotes the Lipschitz constant of $f$ with respect to the metric on $\mathcal{E}$), then $\ipm$ is the Wasserstein or the Kantorovich distance. The Minkowski function for the Wasserstein distance is $\rho_{\mathfrak{F}^W}(f) = L_f$.
                \item[3.] \label{def:mmd-dist}{\bf Maximum Mean Discrepancy (MMD) Distance}: Let $\mathcal{U}$ be a reproducing kernel Hilbert space (RKHS) of real-valued functions on $\mathcal{E}$ and $\mathfrak{F}$ is choosen as $\mathfrak{F}^{MMD} \define \{f\in \mathcal{U}: \Vert f \Vert _{\mathcal{U}} \leq 1 \}$, (where $\Vert \cdot \Vert_{\mathcal{U}}$ denotes the RKHS norm), then $\ipm$ is the Maximum Mean Discrepancy (MMD) distance and its Minkowski functional is $\rho_{\mathfrak{F}^{\text{MMD}}}(f) = \Vert f\Vert _{\mathcal{U}}$.
            \end{compactitem}
 
        % As mentioned previously, we can use the history of state, action and observations to learn the feature abstractions, but it possible to simplify this information structure further. Note that, as the environment is an MDP, the state $\State_\timestep$ is a sufficient statistic. The loss of information happens when $\State_\timestep$ is mapped to $\Feature_\timestep$. This is a subtle yet critical distinction between POMDPs and function approximation in MDPs. In POMDPs the state signal is unavailable, and the system only sees a observation vector $\Feature_\timestep$. Whereas, in function approximation setup, we can observe the state signal perfectly and partial observability is induced due to feature abstraction. Therefore, we can discard the the history of states and focus on the policies of the following form $\policy{}_\timestep: \historyspace_{\timestep} \to \Delta(\actionspace)$ where, $\historyspace_\timestep \define \{\statespace,\featurespace^{1:\timestep-1}, \actionspace^{1:\timestep-1}\}$. Despite this simplification the size of history is growing with time and history compression is still required. As such, we will now proceed towards defining the AIS for MDPs to address all of the aforementioned issues. 
        
        \subsection{Approximate information state}
            
            Given an MDP $\mdp$ and a feature space $\aisspace$, let $\historyspace_\timestep = \statespace \times  \actionspace $ denote the space of all histories $(\State_{1:\timestep}, \Action_{1:\timestep-1})$ up to time~$t$, where $\State_{1:\timestep}$ is a shorthand notation for the history of states $(\State_1,\ldots, \State_\timestep)$, and similar interpretation holds for $\Action_{1:\timestep}$. We are interested in learning history-based feature abstraction functions $\{ \aisfunction_\timestep \colon \historyspace_\timestep \to \aisspace \}_{\timestep \ge 1}$ and a time homogenous policy $\mu \colon \aisspace \to \Delta(\actionspace)$ such that the flattened policy $\policy{} = \{\policy{}_\timestep\}_{\timestep \ge 1}$, where $\policy{}_\timestep = \mu \circ \aisfunction_\timestep$, is approximately optimal.
            
            % We are interested in practical history-based feature abstractions, which often tend to be implemented as RNNs. So, we consider abstractions which are recursively updatable as defined below:
        
            Since the feature abstraction approximates the state, its quality depends on how well it can be used to approximate the per step reward and predict the next state. We formalise this intuition in definition below.

            \begin{definition}\label{def:state-update}
                A family of history-based feature abstraction functions  $\{\aisfunction_\timestep: \historyspace_{\timestep} \to \featurespace\}_{\timestep \geq 1}$ are said to be \emph{recursively updatable} if there exists an update function $\hat f: \featurespace \times \statespace \times \actionspace \to \featurespace $ such that the process $\{\Feature_\timestep\}_{\timestep\geq 1}$, where $\Feature_\timestep = \aisfunction_\timestep(\State_{1:\timestep}, \Action_{1:\timestep-1})$, satisfies:
                \begin{equation}
                    \Feature_{\timestep +1} = \hat f(\Feature_\timestep, \State_{\timestep+1}, \Action_\timestep).\quad \timestep \geq 1
                \end{equation}
            \end{definition}
            \begin{definition}\label{def:ais}
                Given a family of history based recursively updatable feature abstraction functions $\{\aisfunction_\timestep: \historyspace_{\timestep} \to \featurespace\}_{\timestep \geq 1}$, the features $\Feature_\timestep = \aisfunction_\timestep(\State_{1:\timestep}, \Action_{1:\timestep-1})$ are said to be \emph{$(\epsilon, \delta)$-approximate information state} (AIS) with respect to a function space $\mathfrak{F}$ if there exist: (i)~a reward approximation function $\hat r: \featurespace \times \actionspace \to \real$, and (ii)~an approximate transition kernel $\hat\transition: \featurespace \times \actionspace \to \Delta(\statespace)$ such that $\Ais$ satisfies the following properties:
            \begin{compactitem}
                \item[(P1)] \label{p1} Sufficient for approximate performance evaluation: for all $\timestep$,
                \begin{equation}
                     | \cost(\State_{\timestep}, \Action_{\timestep})
                                            - \hat{\cost}(\Feature_\timestep, \Action_\timestep)| \leq \epsilon.
                 \label{def:p1}
                \end{equation}
                \item[(P2)]\label{p2b} Sufficient for predicting future states approximately: for all $\timestep$
                \begin{equation}
                 d_{\mathfrak{F}}(\transition(\cdot \vert \State_\timestep, \Action_\timestep), \hat \transition(\cdot\vert \Feature_\timestep, \Action_\timestep)) \leq \delta.
                % \text{with }\kappa_{\ipm}(\hat{f}_\timestep) &\define \sup_{\history_\timestep, \action_\timestep} \bigg[ \kappa_{\ipm}(\hat{f}_\timestep (\aisfunction_\timestep (\history_\timestep), \cdot, \action_\timestep))\bigg].
                \end{equation}
            \end{compactitem}
        \end{definition}
        We call the tuple $(\hat r ,\hat\transition)$ as an $(\epsilon, \delta)$-AIS approximator. Note that similar definitions have appeared in other works \eg, latent state \citep{deepmdp}, and approximate information state for for POMDPs \citep{ais-1,ais-2}. However, in \citep{deepmdp} it is assumed that the feature abstractions are memory-less and the discussion is restricted to Wasserstein distance. The key difference from the POMDP model in \citep{ais-1,ais-2} is that the in POMDPs the observation $\Feature_\timestep$ is a pre-specified function of the state while in the proposed model $\Feature_\timestep$ depends on our choice of feature abstraction.
        
        As such, our key insight is that an AIS-approximator of a recursively updatable history-based feature abstraction can be used to define a dynamic program. In particular, given a history-based abstraction function $\{\aisfunction_\timestep: \historyspace_\timestep \to \featurespace\}_{\timestep \geq 1}$ which is recursively updatable using $\hat f$ and an $(\epsilon, \delta)$ AIS-approximator $(\hat \transition, \hat \cost)$, we can define the following dynamic programming decomposition: 
        
        For any $\feature_\timestep \in \featurespace, \ \action_\timestep \in \actionspace$
        \begin{subequations}\label{eq:ais-dp}
            \begin{align}
                   \hat Q(\ais_\timestep, \action_\timestep) = \hat \cost(\ais_\timestep, \action_\timestep) + \discount \sum_{\sts_{\timestep+1} \in \statespace}
                    \hat \transition(\sts_{\timestep+1}|\ais_\timestep,\action_\timestep) \hat \valuefunction(\hat{f}(\ais_\timestep,\sts_{\timestep+1},\action_\timestep));&&
                    \hat \valuefunction(\ais_\timestep) = \max_{\action_\timestep \in \actionspace} \hat Q(\ais_\timestep,\action_\timestep), \quad\forall \feature_\timestep \in \featurespace 
            \end{align}
        \end{subequations}
        \begin{definition}\label{def:policy}
            Define $\mu \colon \aisspace \to \Delta(\actionspace)$ be any policy such that for any $\ais \in \aisspace$,
            \begin{align}
                \support(\mu(\ais)) \subseteq 
                \arg\max_{\action \in \actionspace} \hat Q(\ais,\action).\label{eq:policy}
            \end{align}
         Since $\mu$ is a policy from the feature space to actions, we can use it to define a policy from the history of the state action pairs to actions as:
         \begin{align}
             \policy{}_{\timestep}(\sts_{1:\timestep}, \action_{1:\timestep-1}) \define \mu(\aisfunction_{\timestep}(\sts_{1:\timestep}, \action_{1:\timestep-1})) \label{eq:policy-definition}
         \end{align}
        \end{definition}
        Therefore, the dynamic program defined in \eqref{eq:ais-dp} indirectly defines a history-based policy $\policy{}$. The performance of any such history-based policy is given by the following dynamic program: 
        
        For any $\feature \in \featurespace, \ \action \in \actionspace$
        \begin{subequations}
            \begin{align}
                     Q^{\policy{}}_{\timestep}(\history_\timestep, \action_\timestep) = \cost(\sts_\timestep, \action_\timestep) + \discount \sum_{\sts_{\timestep+1} \in \statespace}
                    \transition(\sts_{\timestep+1}|\sts_\timestep,\action_\timestep) \valuefunction_{\timestep+1}^{\policy{}}(\history_{\timestep+1}); && 
                     \valuefunction_{\timestep}^{\policy{}}(\history_\timestep) = \max_{\action \in \actionspace}  Q_{\timestep}^{\policy{}}(\history_\timestep,\action_\timestep), \label{eq:hist-dp}
            \end{align}
        \end{subequations}
        We want to quantify the loss in performance when using the history based policy $\policy$. Note that since $\valuefunction_\timestep^\policy$ is not time-homogeneous, we need to compute the worst-case difference between $\valuefunction^{\star}$ and $\valuefunction_\timestep^\policy$, which is given by:
        \begin{equation}
            \Delta \define\sup_{\timestep \geq 0}\sup_{\history_\timestep = (\sts_{1:\timestep}, \action_{1:\timestep}) \in \historyspace_\timestep} \vert \valuefunction^{\star}(\sts_\timestep) - \valuefunction^{\policy{}}_{\timestep}(\history_\timestep)\vert, \label{eq:sup-v}
        \end{equation}
        
        Our main approximation result is the following:
    
        % A natural question which then arises is that \emph{how far from optimal is $\policy{}$?}
        % We answer this question in the following result:
        \begin{theorem}\label{thm:ais-dp}
            % For any time $\timestep$, any realisation $\sts_\timestep$ of $\State_\timestep$, $\action_\timestep$ of $\Action_\timestep$, let $\history_\timestep = (\sts_{1:\timestep}, \action_{1:\timestep-1})$, and $\ais_\timestep = \aisfunction_\timestep(\history_\timestep)$.
             The worst case difference between $\valuefunction^{\star}$ and $\valuefunction^{\policy{}}_{\timestep}$ is bounded by
            %     \begin{align}
            %     \Delta &\define \sup_{\timestep \geq 0}\sup_{\history_\timestep = (\sts_{1:\timestep}, \action_{1:\timestep}) \in \historyspace_\timestep} \vert \valuefunction^{\star}(\sts_\timestep) - 
            %      \valuefunction^{\policy{}}(\ais_\timestep)\vert. \label{eq:sup-v}
            %     \end{align}
            % Then 
            \begin{equation}
                \Delta
                \le 2 \frac{\varepsilon + \discount\delta \kappa_{\mathfrak{F}}(\hat \valuefunction^{\mu}, \hat{f})}{1 - \discount},\label{eq:ais-bound}
            \end{equation}
        where $\kappa_{\mathfrak{F}}(\hat \valuefunction^{\mu}, \hat f)$ = $ \sup_{\feature, \action}\rho_{\mathfrak{F}}(\hat \valuefunction^{\mu}(\hat f(\cdot, \feature, \action)))$, $\rho_{\mathfrak{F}}(\cdot)$ is the Minkowski functional associated with the IPM $\ipm$ as defined in \eqref{eq:minkowski-functional}.
        \end{theorem}
        Proof in \Cref{sec:proof:thm:ais-dp}

        Some salient features of the bound are as follows:
        First, the bound depends on the choice of metric on probability spaces. Different IPMs will result in a different value of $\delta$ and also a different value of $\kappa_{\mathfrak F}(\hat{\valuefunction}^{\mu}, \hat f)$. Second, the bound depends on the properties of $\hat{\valuefunction}^{\mu}$. For this reason we call it an instance dependent bound. Sometimes, it is desirable to have bounds which do not require solving the dynamic program in \eqref{eq:ais-dp}. We present such bounds as below, note that these ``instance independent'' bounds are the derived by upper bounding $\kappa_{\mathfrak{F}}(\hat{\valuefunction}^{\mu}, \hat f)$. Therefore, these are looser than the upper bound in \Cref{thm:ais-dp}
        
        % In \eqref{eq:ais-bound}, $\epsilon$ and $\delta$, capture the worst case error incurred by the AIS when predicting instantaneous reward $\cost$ and approximating the transition distribution of the ground MDP $\mdp$. Therefore, resulting suboptimality bound helps us quantify us the loss in performance due to feature abstraction. At the same time, the IPM used for measuring the distance between the approximate and true transition distribution also influences the bound via $\kappa_{ \mathfrak{F}}(\hat \valuefunction^{\mu}, \hat f)$. In the following corollaries we will show how $\kappa_{ \mathfrak{F}}(\hat \valuefunction^{\mu}, \hat f)$ takes a specific form according the choice of the IPM.
        
         \begin{corollary}\label{THM:TV-BOUND}
             If the function class $\mathfrak{F}$ is $\mathfrak{F}^{\text{TV}}$, then $\Delta$ as defined in \eqref{eq:sup-v} is upper bounded as:
             \begin{align}
              \Delta
                \le  \frac{2\epsilon}{(1-\discount)} +  \frac{\discount\delta \spn(\hat\cost)} {(1-\discount)^2}.
            \end{align}
        \end{corollary}
        
            Proof in Appendix \ref{sec:tv-proof}
        
        \begin{corollary}\label{THM:LIP-BOUND}
            Let $L_{\hat \cost}$ and $L_{\hat{\transition}}$ denote the Lipschitz constants of the approximate reward function $\hat \cost$ and approximate transition function $\hat \transition$ respectively, and $L_{\hat f}$ is the uniform bound on the Lipschitz constant of $\hat f$ with respect to the state $\State_\timestep$.
            If $\discount L_{\hat \transition}L_{\hat f} \leq 1$ and the function class $\mathfrak{F}$ is $\mathfrak{F}^{\text{W}}$, then $\Delta$ as defined in  \eqref{eq:sup-v} is upper bounded as:
            \begin{align}
                \Delta
                \le  \frac{2\epsilon}{(1-\discount)} + \frac{2\discount\delta L_{\hat \cost} }{(1- \discount)(1-\discount L_{\hat f}L_{\hat\transition})}.
            \end{align}
        \end{corollary}

            Proof in Appendix \ref{sec:w-proof}

        \begin{corollary}\label{thm:mmd-bound}
             If the function class $\mathfrak{F}$ is $\mathfrak{F}^{\text{MMD}}$, then $\Delta$ as defined in \eqref{eq:sup-v} is upper bounded as:
             \begin{align}
              \Delta
                \le  2 \frac{\epsilon + \discount\delta\kappa_{\mathcal{U}}(\hat \valuefunction, \hat f) } {(1-\discount)},
            \end{align}
            where $\mathcal{U}$ is a RKHS space,  $\Vert\cdot\Vert_{\mathcal{U}}$ its associated norm and $\kappa_{\mathcal{U}}(\hat \valuefunction, \hat f) = \sup_{\feature, \action}\Vert(\hat \valuefunction(\hat f(\cdot, \feature, \action)))\Vert_{\mathcal{U}}$.
        \end{corollary}
        \begin{proof}
            The proof follows from the properties of MMD described previously. 
        \end{proof}
        
        In the following section we will show how one can use these theoretical insights to design a policy search algorithm.
        
\section{Reinforcement learning with history-based feature abstraction}\label{sec:algorithm}
In this section, we leverage the approximation bounds of
Theorem~\ref{thm:ais-dp} to develop a reinforcement learning algorithm. The
main idea is to add an additional block, which we call the AIS-approximator,
to any standard RL algorithm. In this section, we explain an AIS-based
generalization for policy-based algorithms such as REINFORCE and actor-critic, but the same idea could be used for
value-based algorithms such as Q-learning as well. 
% \begin{figure*}
%     \centering
%     \includegraphics[width=\linewidth]{AAMAS-2023-Formatting-Instructions/Results/nns.pdf}
%     \caption{AIS approximator block} \label{fig:blk-diag}
% \end{figure*}

\begin{wrapfigure}{rt}{0.6\textwidth}
      \vspace*{-6mm}
      \includegraphics[width=0.6\textwidth]{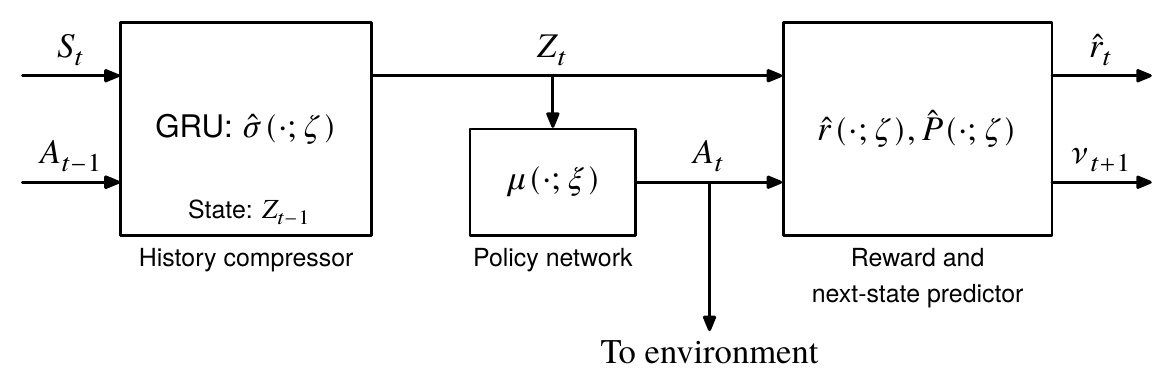}
      \vspace*{-5mm}
      \caption{AIS approximator block} \label{fig:blk-diag}
      \vspace*{-5mm}
\end{wrapfigure}

The AIS-approximator consists of two blocks: a recursively updatable history
compressor and a reward and next-state predictor as shown in
Fig.~\ref{fig:blk-diag}. In particular, we can consider any parameterised family of
the history compression functions
$\{\aisfunction_\timestep(\cdot); \aisparams) \colon \historyspace_\timestep \to
\aisspace\}$ which are recursively updatable via the function
$\hat{f}(\cdot) \colon \aisspace \times \statespace\times \actionspace \to
\aisspace$ as the history-compressor along with any parameterised family of
functions $\hat \cost(\cdot; \aisparams)\colon
\aisspace \times \actionspace \to \real$ as the reward approximator and any
parameterised stochastic kernels ${\hat
\transition}(\cdot;\aisparams)\colon\aisspace \times \actionspace \to
\Delta(\statespace)$ as the transition approximator. In the above notation $\aisparams$ denotes the
combined parameters of the family of functions. As a concrete example, we could use 
use memory-based neural networks such as LSTMs or GRUs as the
history-compression functions. The memory update functions of such networks
correspond to the update function $\hat f$. A multilayered perceptron (MLP)
could be used as a reward approximator and a parameterized family of
stochastic kernels such as the softmax function or a mixture of Gaussians
could be used as the transition approximator. The parameters of all these
networks together are denoted by $\aisparams$.

We use a weighted combination of the reward prediction loss $\vert
\cost(\State_\timestep, \Action_\timestep) - \hat\cost (\Ais_\timestep,
\Action_\timestep)\vert$ and the transition-prediction loss $\ipm(\transition,
\hat \transition)$ as the loss function for the AIS-generator. In particular,
the AIS-loss is given by
    \begin{align}
          \aisloss(\aisparams) &= \frac{1}{\Timestep}\sum_{t = 0}^{\Timestep}\bigg( \lambda \underbrace{({\hat{\cost}}(\Ais_{\timestep}, \Action_\timestep; \aisparams)  - \cost(\State_\timestep, \Action_\timestep))^{2}}_{\loss_{\hat{\Cost}(\cdot;\aisparams)}}+ (1-\lambda)\cdot \underbrace{\ipm({\hat\transition}(\Ais_\timestep, \Action_\timestep\ ;\aisparams),\transition)^{2}}_{\loss_{\hat\transition}(\cdot;\aisparams)}\bigg),\label{eq:pgt-loss}
    \end{align}
    where $\Timestep$ is the length of the episode or the rollout length, $\lambda \in [0,1]$ is a hyper-parameter.
    % reward prediction loss $\loss_{\hat{\Cost}}(;\aisparams)$ is simply the mean-squared error between the predicted and the observed reward, whereas the transition prediction loss $\loss_{\hat{\transition}}(\cdot; \aisparams)$ is the distance between predicted and observed transition distributions $\hat\transition$ and $\transition$. 
    The computation of $\loss_{\hat{\transition}}(\cdot; \aisparams)$, depends on the choice of IPM. In principle we can pick any IPM, but we would want to use an IPM using which the distance $d_{\mathfrak{F}}$ can be efficiently computed.

    \subsection{Choice of an IPM} \label{sec:ipm-choice}

        To compute the IPM $d_\mathfrak{F}$ we need to know the probability density functions $\hat \transition$ and $\transition$. As we assume $\hat \transition$ to belongs to a parametric family, we know its density function in closed form. However, since we are in the learning setup, we can only access samples from $\transition $. For a function a $f\in \mathfrak{F}$, and probability density functions $\transition$ and $\hat \transition$ such that, $\nu_1 = \transition$, and $\nu_2 =\hat \transition$, we can estimate the IPM $d_\mathfrak{F}$ between a distribution and samples using the duality $|\int_\aisspace f d\nu_1 - \int_\aisspace f d\nu_2|$. In this paper, we use two from of IPMs, the MMD distance and the Wasserstein/Kantorovich–Rubinstein distance. 
        
        \subsubsection{MMD Distance:} Let $m_\aisparams$ denote the mean of the distribution $\hat \transition(\cdot;\aisparams)$. Then, the AIS-loss when MMD is used as an IPM is given by
        %Of the two alternatives, MMD distance is the easier compute as we can use some of its properties to simply its computation. Towards that end, when $\hat \transition$ is a real-valued distribution that can be characterised by its mean $m_\aisparams$, the MMD-based AIS loss can be given as:
        \begin{align}
           \aisloss(\aisparams) &= \frac{1}{\Timestep}\sum_{t = 0}^{\Timestep}\bigg( \lambda ({\hat{\cost}}(\Ais_{\timestep}, \Action_\timestep; \aisparams)  - \cost(\State_\timestep, \Action_\timestep))^{2} + (1-\lambda)(m^{\State_\timestep}_{\aisparams} - 2\State_\timestep)^{\top}m^{\State_\timestep}_{\aisparams}  \bigg),\label{eq:mmd-ais-loss}
        \end{align}
        where $m^{\State_\timestep}_{\aisparams}$ is obtained using the from the transition approximator, ~\ie, the mapping ${\hat\transition}(\aisparams): \aisspace \times \actionspace \to \real$. For the detailed derivation of the above loss see \Cref{sec:mmd-details}
        
        \subsubsection{Wasserstein/Kantorovich–Rubinstein distance:} 
        In principle, the Wasserstein/Kantorovich distance can be computed by solving a linear program~\citep{Sriperumbudur}, but doing at every episode can be computationally expensive. 
        %When $\ipm$ is the Wasserstein/Kantorovich–Rubinstein distance, we can compute it by solving a linear program~\citep{Sriperumbudur}. In some settings, it might be computationally expensive to solve a linear program at each time step before updating the parameters $\aisparams$. 
        Therefore, we propose to approximate the Wasserstein distance using a KL-divergence~\citep{kl} based upper-bound. The simplified-KL divergence based AIS loss is given as:
        \begin{align}
                \aisloss(\aisparams) &= \frac{1}{\Timestep} \sum_{t = 0}^{\Timestep}\bigg( \lambda ({\hat{\cost}}(\Ais_{\timestep}, \Action_\timestep; \aisparams)  - \cost(\State_\timestep, \Action_\timestep))^{2} + (1-\lambda)\log(\hat \transition(\State_\timestep;\aisparams))  \bigg),\label{eq:w-ais-loss}
        \end{align} 
        where after dropping the terms which do not depend on $\aisparams$, we get $d_{\mathfrak{F}^{\text{W}}}^{2}(\transition, \hat \transition)\leq \log(\hat \transition(\State_\timestep;\aisparams))$ is the simplified-KL-divergence based upper bound. For the details of this derivation see \Cref{sec:wass-details}.

   \subsection{Policy gradient algorithm}\label{sec:pgt}
    % \begin{minipage}{\linewidth}
        \begin{algorithm}
            \SetKwData{Left}{left}\SetKwData{This}{this}\SetKwData{Up}{up}
            \SetKwFunction{Union}{Union}\SetKwFunction{FindCompress}{FindCompress}
            \SetKwInOut{Input}{Input}\SetKwInOut{Output}{Output}
            \SetAlgoLined
            \Input{$\iota_{0}$: Initial state distribution, \\
                  $\aisparams_{0}$: Ais parameters, \\
                  $\actorparams_{0}$: Actor parameters, \\
                %   $\criticparams_{0}$: Critic parameters,
                %   $\ais_{0}$: Initial Ais, 
                  $\action_{0}$: Initial action, \\
                  $\mathcal{D} = \emptyset$: Replay buffer, \\
                  $N_{\text{comp}}$: Computation \rlap{budget,} \\
                  $N_{\text{ep}}$: Episode length, \\
                  $N_{\text{grad}}$: Gradient steps}
            % \KwResult{Write here the result }
             \For{iterations $i = 0:N_{\text{comp}}$ }{
                  Sample start state $\sts_{0}\sim \iota_{0}$\;
                  \For{iterations $j = 0:N_{\text{ep}}$ }
                  {
                  $\ais_{j} = \aisfunction_{\aisparams}(\sts_{1:j},\action_{1:j-1})$\;
                  % \tcp*{\tiny{final layer of a GRU cell.}}
                  $\action_{j} = \mu_{\actorparams}(\ais_{j})$\;
                  $\sts_{j+1} = \transition(\sts_{j},\action_{j})$\; %\tcp*{\tiny{sample next state from the MDP.}}
                  $\mathcal{D} \xleftarrow{} \{\ais_{j},\action_{j},\sts_{j},\sts_{j+1}\}$\;
                  $\action_{j-1} = \action_{j}$\;
                  $\sts_{j} = \sts_{j+1}$\;
                  }
                  \For{every batch $b \in \mathcal{D}$}
                  {
                        \For {gradient step $t=0:N_{\text{grad}}$}
                        {
                        %  $\aisparams_{\timestep+1,b,\valuefunction} = \aisparams_{\timestep,b,\valuefunction} + \aislr  \grad_{\aisparams_{\valuefunction}}\aisloss(\aisparams_{\timestep,b,\valuefunction})$\;
                        %  $\criticparams_{\timestep+1,b} = \criticparams_{\timestep,b} + \criticlr\grad_{\criticparams}\criticloss(\criticparams_{\timestep,b})$\;
                         $\aisparams_{\timestep+1,b} = \aisparams_{\timestep,b} + \aislr \grad_{\aisparams}\aisloss(\aisparams_{\timestep,b})$\;
                         $\actorparams_{\timestep+1,b} = \actorparams_{\timestep,b} +\actorlr \hat\grad_{\actorparams}\performance(\actorparams_{\timestep,b},\aisparams_{\timestep,b})$
                        }
                  }
             }
          \caption{Policy Search with \rlap{AIS}}\label{alg:ciac-a}
      \end{algorithm}

        Following the design of the AIS block, we now provide a policy-gradient algorithm to learning both the AIS and policy. The schematic of our agent architecture is given in \Cref{fig:blk-diag}, and pseudo-code is given in \Cref{alg:ciac-a}. Given a feature space $\aisspace$, we can simultaneously learn the AIS-generator and the policy using a multi-timescale stochastic gradient ascent algorithm~\citep{borkar2008stochastic}. Let $\mu(\cdot;\actorparams):\aisspace \to \Delta(\actionspace)$ be a parameterised stochastic policy with parameters $\actorparams$. Let $\performance(\actorparams,\aisparams)$ denote the performance of the policy $\mu(\cdot ;\ \actorparams)$. The policy gradient theorem~\citep{pgt,Williams2004SimpleSG,baxter-bartlett} states that: 
        % \begin{align*}
        %     \grad_{\actorparams}\performance(\actorparams_\timestep,\aisparams_\timestep) &= \expecun{}\bigg[\sum_{\timestep=1}^{\infty}\discount^{\timestep-1}\cost_{\timestep}\bigg(\sum_{\tau =1}^{\timestep}\grad_{\actorparams}\log(\mu(\Action_\timestep|\Ais_\timestep ;\ \actorparams_\timestep))\bigg)\bigg].\label{eq:pg}
        % \end{align*}
        For a rollout horizon $\Timestep$, we can estimate $\grad_\actorparams \performance$ as:
        \begin{align*}
            \hat \grad_{\actorparams}\performance(\actorparams_\timestep,\aisparams_\timestep) &= \sum_{\timestep=1}^{\Timestep}\discount^{\timestep-1}\cost_{\timestep}\bigg(\sum_{\tau =1}^{\timestep}\grad_{\actorparams}\log(\mu(\Action_\timestep|\Ais_\timestep ;\ \actorparams_\timestep))\bigg).
        \end{align*}
        Following a rollout of length $\Timestep$, we can then update the parameters $\{(\aisparams_i, \actorparams_i  )\}_{i \geq 1}$ as follows:
        
        \begin{subequations}\label{eq:pgt-update}
             \begin{align}
                \aisparams_{i+1} = \aisparams_i + \aislr_i \grad_\aisparams\aisloss(\aisparams_i), &&
                \actorparams_{i+1} = \actorparams_i + \actorlr_i \hat\grad_{\actorparams}\performance(\actorparams_{i},\aisparams_{i})\label{eq:actor-update},
            \end{align}
        \end{subequations}
            % \begin{align}
            %     \aisparams_{i+1} &= \aisparams_i + \aislr_i \grad_\aisparams\aisloss(\aisparams_i),
            % \label{eq:ais-update}
            % \\
            %     \actorparams_{i+1} &= \actorparams_i + \actorlr_i \hat\grad_{\actorparams}\performance(\actorparams_{i},\aisparams_{i})
            % \label{eq:actor-update}
            % \end{align}
         where the step-size $\{\aislr_{i}\}_{i \geq 0}$ and $\{\actorlr_{i}\}_{i \geq 0}$ satisfy the standard conditions $\sum_{i} \aislr_{i} = \infty$, $\sum_{i}\aislr_{i}^{2}< \infty$, $\sum_{i} \actorlr_{i} = \infty$ and $\sum_{i}\actorlr_{i}^{2}< \infty$ respectively. Moreover, one can ensure that the AIS generator converges faster by choosing an appropriate learning rates such that, $\lim_{i \to \infty} \frac{\actorlr_{i}}{\aislr_{i}} = 0$. 
         
    \subsection{Actor Critic Algorithm} \label{sec:AC}
        We can also use the aforementioned ideas to design an AIS based actor-critic algorithm. In addition to a parameterised policy $\policy(\cdot; \actorparams)$ and AIS generator $(\aisfunction_\timestep(\cdot;\aisparams), \hat f, \hat r, \hat\transition)$ the actor-critic algorithm uses a parameterised critic $\hat\valuefunction(\cdot;\criticparams):\featurespace \to \real$, where $\criticparams$ are the parameters for the critic. The performance of policy $\mu(\cdot;\actorparams)$ is then given by $\performance(\actorparams, \aisparams, \criticparams)$. According to policy gradient theorem~\citep{pgt,baxter-bartlett} the gradient of $\performance(\actorparams, \aisparams, \criticparams)$, is given as:
        \begin{align}
            \grad_\actorparams \performance(\actorparams, \aisparams, \criticparams) &= \expecun{}\bigg[ \grad_{\actorparams}\log(\mu(\cdot;\actorparams))\hat{\valuefunction}(\cdot;\criticparams)\bigg].
        \end{align}
        And for a trajectory of length $\Timestep$, we approximate it as:
        \begin{align}
            \hat{\grad}_\actorparams \performance(\actorparams, \aisparams, \criticparams) &= \frac{1}{\Timestep}\sum_{\timestep =1}^{\Timestep}\bigg[ \grad_{\actorparams}\log(\mu(\cdot;\actorparams))\hat{\valuefunction}(\cdot;\criticparams)\bigg].
        \end{align}
        The parameters $\criticparams$ can be learnt by optimising the temporal difference loss given as:
        \begin{align}
            \loss_{\text{TD}}(\actorparams, \aisparams, \criticparams) &= \frac{1}{\Timestep}\sum_{\timestep=0}^{\Timestep}\texttt{smoothL1}(\hat{\valuefunction}(\Feature_\timestep;\criticparams) - \cost(\Feature_\timestep,\Action_\timestep) - \discount \hat{\valuefunction}(\Feature_{\timestep+1};\criticparams)).
        \end{align}
         The parameters $\{(\aisparams_i, \actorparams_i, \criticparams_i  )\}_{i \geq 1}$ can then be updated using a multi-timescale stochastic approximation algorithm as follows:
         \begin{subequations}\label{eq:ac-update}
            \begin{align}
                \aisparams_{i+1} &= \aisparams_i + \aislr_i \grad_\aisparams\aisloss(\aisparams_i)\label{eq:ac-ais-update}\\
                \criticparams_{i+1} &= \criticparams_i + \criticlr_i \grad_{\criticparams}\loss_{\text{TD}}(\actorparams_i, \aisparams_i, \criticparams_i)\label{eq:ac-critic-update}\\
                \actorparams_{i+1} &= \actorparams_i + \actorlr_i \hat{\grad}_{\actorparams}\performance(\actorparams_{i},\aisparams_{i},\criticparams)\label{eq:ac-actor-update},
            \end{align}
         \end{subequations}
          where the step-size $\{\aislr_{i}\}_{i \geq 0}$,  $\{\criticlr_{i}\}_{i \geq 0}$ and $\{\actorlr_{i}\}_{i \geq 0}$ satisfy the standard conditions $\sum_{i} \aislr_{i} = \infty$, $\sum_{i}\aislr_{i}^{2} < \infty$, $\sum_{i} \criticlr_{i} = \infty$, $\sum_{i}\criticlr_{i}^{2} < \infty$, $\sum_{i} \actorlr_{i} = \infty$ and $\sum_{i}\actorlr_{i}^{2}< \infty$ respectively. Moreover, one can ensure that the AIS generator converges first, followed by the critic and the actor by choosing an appropriate step-sizes such that, $\lim_{i \to \infty} \frac{\actorlr_{i}}{\aislr_{i}} = 0$ and $\lim_{i \to \infty} \frac{\criticlr_{i}}{\actorlr_{i}} = 0$.

        %  Note that we can use similar ideas to develop an Actor-Critic algorithm where, in addition to a parameterised policy $\policy(\cdot; \actorparams)$ and AIS generator $(\aisfunction_\timestep(\cdot;\aisparams), \hat f, \hat r, \hat\transition)$ we can also a parameterised critic $\hat\valuefunction(\cdot;\criticparams):\featurespace \to \real$, where $\criticparams$ are the parameters for the critic. The details of the Actor Critic Algorithm can be found in \Cref{sec:AC}, and 
        % The convergence analysis of both algorithms can be found in \Cref{sec:convergence}.
        
         \subsection{Convergence analysis}\label{sec:convergence}
    
    In this section we will discuss the convergence of the AIS-based policy gradient in \Cref{sec:pgt} as well as Actor-Critic algorithm presented in the previous subsection. The proof of convergence relies on multi-timescale stochastic approximation \citet{borkar2008stochastic} under conditions similar to the standard conditions for convergence of policy gradient algorithms with function approximation stated below, therefore it would suffice to provide a proof sketch.

    \noindent\begin{assumption} \label{assumption-1}
        \begin{enumerate}
            \item \label{a.1.1}The values of step-size parameters $\aislr, \actorlr$ and $\criticlr$ (for the actor critic algorithm) are set such that the timescales of the updates for $\aisparams$, $\actorparams$, and $\criticparams$ (for Actor-Critic algorithm) are separated, ~\ie, $\aislr_{\timestep} \gg \actorlr_{\timestep}$, and for the Actor-Critic algorithm $\aislr_{\timestep} \gg \criticlr_\timestep \gg \actorlr_{\timestep}$, $\sum_{i} \aislr_{i} = \infty$, $\sum_{i}\aislr_{i}^{2} < \infty$, $\sum_{i} \criticlr_{i} = \infty$, $\sum_{i}\criticlr_{i}^{2} < \infty$, $\sum_{i} \actorlr_{i} = \infty$ and $\sum_{i}\actorlr_{i}^{2}< \infty$, $\lim_{i \to \infty} \frac{\actorlr_{i}}{\aislr_{i}} = 0$ and $\lim_{i \to \infty} \frac{\criticlr_{i}}{\actorlr_{i}} = 0$, 
            \item \label{a.1.2}The parameters $\aisparams$, $\actorparams$ and $\criticparams$ (for Actor-Critic algorithm) lie in a convex, compact and closed subset of Euclidean spaces.
            \item \label{a.1.3}The gradient $\grad_{\aisparams}\aisloss$ is Lipschitz in $\aisparams_{\timestep}$, and $\hat \grad_{\actorparams}\performance(\actorparams,\aisparams)$ is Lipschitz in $\actorparams_{\timestep},~\text{and}~\aisparams_{\timestep}$. Whereas for the Actor-Critic algorithm the gradient of the TD loss $ \grad_{\criticparams}\loss_{\text{TD}}(\aisparams, \actorparams, \criticparams)$ and the policy gradient $\hat \grad_{\actorparams} \performance(\aisparams, \actorparams, \criticparams)$ is Lipschitz in $(\aisparams_\timestep, \actorparams_\timestep, \criticparams_\timestep)$.
            \item \label{a.1.4}Estimates of gradients $\grad_{\aisparams}\aisloss$, $\grad_{\actorparams}\performance(\actorparams,\aisparams)$, and $ \grad_{\criticparams}\loss_{\text{TD}}(\aisparams, \actorparams, \criticparams)$ and are unbiased with bounded variance\footnote{This assumption is only satisfied in tabular MDPs.}.
            % Moreover, in the case of the Actor-Critic algorithm, the Critic and the function approximator are compatible as given in \citet{...} \ie,
            % \begin{align*}
            %     \frac{\partial\hat Q_{\criticparams_\timestep}(\Ais_\timestep, \Action_\timestep)}{\partial \criticparams} =\frac{1}{\policy_{\actorparams_{\timestep}}}\frac{\partial \policy_{\actorparams_\timestep}}{\partial \actorparams}.
            % \end{align*}
        \end{enumerate}
    \end{assumption}

        \begin{assumption} \label{assumption-2}
            \begin{enumerate}
                \item \label{a.2.1}The ordinary differential equation (ODE) corresponding to \eqref{eq:actor-update} is locally asymptotically stable.
                \item \label{a.2.2}The ODEs corresponding to \eqref{eq:pgt-update} is globally asymptotically stable.
                \item For the Actor-Critic algorithm, the ODE corresponding to \eqref{eq:ac-critic-update} is globally asymptotically stable and has a fixed point which is Lipschitz in $\actorparams$.
            \end{enumerate}
        \end{assumption}
        \begin{theorem}\label{thm:convergence}
        Under \cref{assumption-1,assumption-2}, along any sample path, almost surely we have the following:
        \begin{enumerate}
            \item The iteration for $\aisparams$ in \eqref{eq:pgt-update} converges to an AIS generator that minimises the $\aisloss$.
            \item The iteration for $\actorparams$ in \eqref{eq:actor-update} converges to a local maximum of the performance $\performance(\aisparams^\star,\actorparams)$ where $\aisparams^\star$, and $\criticparams^\star$ (for Actor Critic) are the converged value of $\aisparams$, $\criticparams$.
            \item For the Actor-Critic algorithm the iteration for $\criticparams$ in \eqref{eq:ac-critic-update} converges to critic that minimises the error with respect to the true value function.
        \end{enumerate}
        \end{theorem}
        \begin{proof}
        % On satisfying \cref{assumption-1}.1, a suitable continuous time interpolation of \eqref{def:ais-update} will be an asymptotic pseudo-trajectory of the semi-flow induced by it's ordinary differential equation (ODE). Therefore, this interpolation will converge to the limit point of \eqref{def:ais-update}'s ODE, and by principle of superposition the ODE will be globally asymptotically stable. As such, by the arguments made by~\citet{borkar2008stochastic,Kushner1997StochasticAA}, iteration given by \eqref{def:ais-update} will converge to its corresponding fixed point. By similar a argument, continuous time interpolations of \eqref{def:actor-update-2} will also converge to the limit points of its respective ODEs.
        The proof for this theorem follows the technique used in \citep{Leslie2004ReinforcementLI,borkar2008stochastic}. Due to the specific choice of learning rate the AIS-generator is updated at a faster time-scale than the actor, therefore it is ``quasi static'' with respect to to the actor while the actor observes a ``nearly equilibriated'' AIS generator. Similarly in the case of the Actor-Critic algorithm the AIS generator observes a stationary critic and actor, whereas the critic and actor see ``nearly equilibriated'' AIS generator. The Martingale difference condition (A3) of \citet{borkar2008stochastic} is satisfied due to \cref{a.1.4} in \cref{assumption-1}. As such since our algorithm satisfies all the four conditions by \citep[page35]{Leslie2004ReinforcementLI}, \citep[Theorem 23]{Borkar1997StochasticAW}, the result then follows by combining the theorem on \citep[page 35]{Leslie2004ReinforcementLI}\citep[Theorem 23]{borkar2008stochastic} and \citep[Theorem 2.2]{Borkar1997StochasticAW}.
        \end{proof}

    \begin{figure*}[!htbp]
      \includegraphics[width=\linewidth]{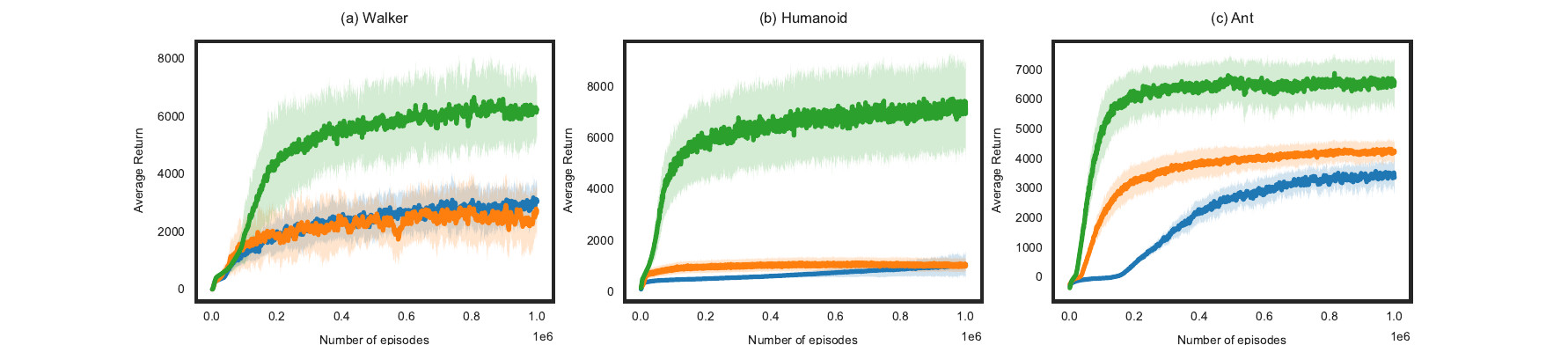}
      % \caption{This is a figure} \label{fig:comb-results-1}
      \includegraphics[width=\linewidth]{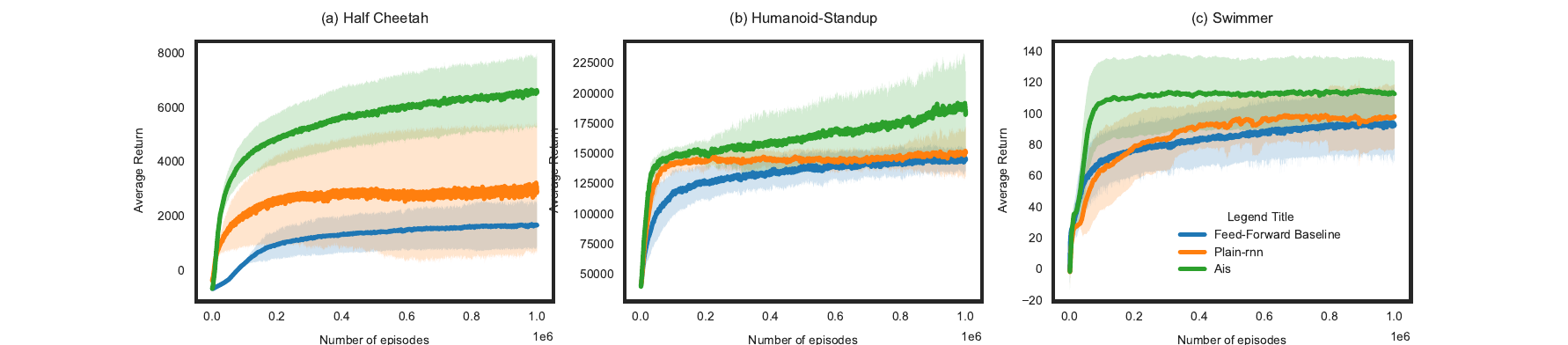}
      \caption{Empirical results averaged over 50 Monte Carlo runs with shaded regions showing the interquantile range.} \label{fig:comb-results-2}
    \end{figure*}

\section{Empirical evaluation}\label{sec:experiments}

    Through our experiments, we seek to answer the following questions:
    (1) Can history-based feature representation policies help improve the quality solution found by a memory-less RL algorithm?
    (2) In regards to the solution quality and sample complexity, how does the proposed method compare with other memory-augmented policies?
    (3) How does the choice of IPM affect the algorithms performance?

     We answer question (1) and (2) by comparing our approach with the proximal policy gradient (PPO) algorithm which uses feed-forward neural networks. For question (2), we compare our method with an LSTM-based PPO variant which learns the feature representation using the history of states $\State_{1:\Timestep}$ in a trajectory. For question (3) we compare the performance of our method using different MMD kernels and KL-divergence based approximation of Wasserstein distance. All the approaches are evaluated on six continuous control tasks from the MuJoCo~\citep{Todorov2012MuJoCoAP} OpenAI-Gym suite. To ensure a fair comparison, the baselines and their respective hyper-parameter settings are taken from well tested stand-alone implementations provided by~\citet{baselines}. From an implementation perspective, our framework can be used to modify any off-the-shelf policy-gradient algorithm by simply replacing (or augmenting) the feature abstraction layers of the policy and/or value networks with recurrent neural networks (RNNs), trained with the appropriate losses, as outlined previously. In these experiments, we replace the fully connected layers in PPO's architecture with a Gated Recurrent Unit (GRU). For all the implementations, we initialise the hidden state of the GRU to zero at the beginning of the trajectory. This strategy simplifies the implementation and also allows for independent decorrelated sampling of sequences, therefore ensuring robust optimisation of the networks~\citep{rnn-hausknecht}. It is important to note that we can extend our framework to other policy gradient methods such as SAC~\citep{HaarnojaZAL18}, TD3~\citep{td3} or DDPG~\citep{ddpg}, after satisfying certain technical conditions. However, we leave these extensions for future work. Additional experimental details and results can be found in \Cref{sec:experiment-details}.

%\subsection{Numerical Results} 
    
     \Cref{fig:comb-results-2} contains the results of our experiments averaged over 50 Monte-Carlo evaluations using MMD-based AIS loss in \eqref{eq:mmd-ais-loss}.  These results show that our algorithm improves over the performance of both the baselines, and the performance gain is significantly higher for high-dimensional environments like Humanoid and Ant. 
     It is worth noticing that the GRU baseline also outperforms the feed-forward baseline for most  environments. Overall, these findings lend credence to  history-based encoding policies as a way to improve the quality of the solution learnt by the RL algorithm.

     \begin{figure*}[!htbp]
        \includegraphics[width=\linewidth]{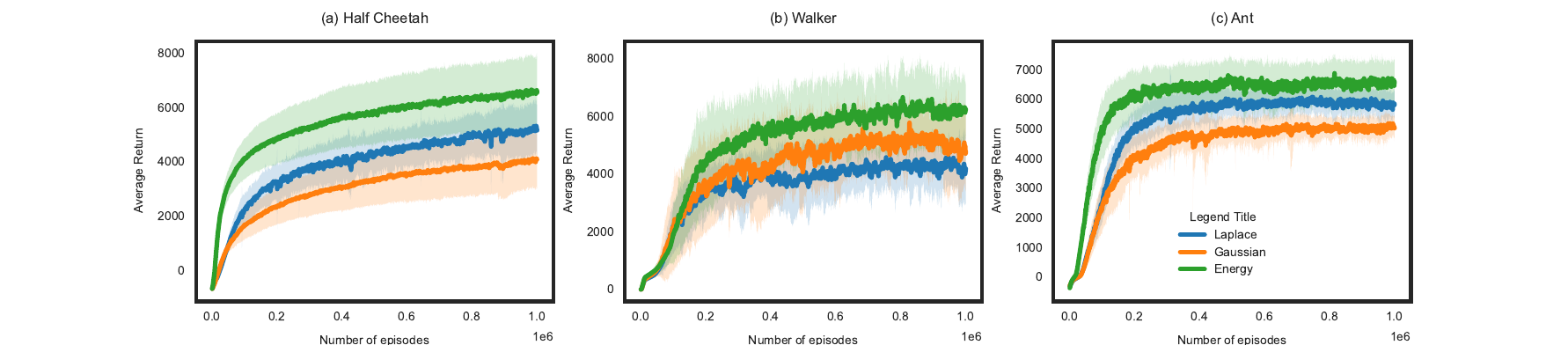}
        \caption{Comparison of different MMDs, averaged over 50 runs} \label{fig:MMD-comp}
    \end{figure*}

    Note that the MMD distance given by \eqref{eq:mmd-grad} in \Cref{sec:mmd-details}, can be computed using different types of characteristic kernels (for a detailed review see ~\citep{Sriperumbudur,NIPS2009_685ac8ca,sejdinovic}). In this paper we consider computing \eqref{eq:mmd-grad} using the Laplace, Gaussian and energy distance kernels. In in \Cref{fig:MMD-comp} we compre the perfromance of our methods under different MMD kernels. It can be observed that for the continuous control tasks in the MuJoCo suite, the energy distance yields better performance, and therefore we implement \cref{eq:mmd-grad} using the energy distance for the results in \Cref{fig:comb-results-2}.
    
\begin{figure*}[!htbp]
            \includegraphics[width=\linewidth]{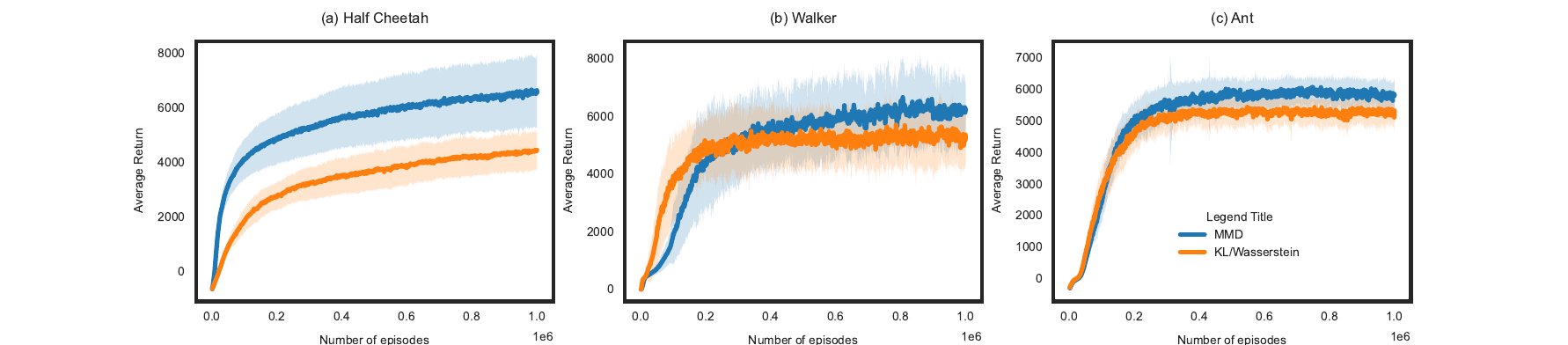}
            \caption{Comparison of Wasserstein vs MMDs, averaged over 50 runs.} \label{fig:Wass-res}
        \end{figure*}

    Next, we compare the performance of our method under MMD (Energy distance kernel) and Wasserstein distance. From \Cref{fig:Wass-res} we observe that for continuous control tasks, use of MMDs result in better performance as compared to Wasserstein distance. 
    
    % \begin{figure}[!htbp]
    %         \includegraphics[width=\linewidth]{Results/dist-new-1.pdf}
    %         \caption{MMD vs KL for Half Cheetah} \label{fig:dist-1}
    % \end{figure}
    
    % \begin{figure}[!htbp]
    %         \includegraphics[width=\linewidth]{Results/dist-new-2.pdf}
    %         \caption{MMD vs KL for Ant } \label{fig:dist-2}
    % \end{figure}
    
    % \begin{figure}[!htbp]
    %         \includegraphics[width=\linewidth]{Results/dist-new-3.pdf}
    %         \caption{MMD vs KL for Walker} \label{fig:dist-3}
    % \end{figure}

\section{Related Work}\label{sec:litreview}

    The development of RL algorithms with memory-based feature abstractions has been an active area of research, and most existing algorithms have tackled this problem using non-parametric methods like Nearest neighbour~\citep{Bentley1975MultidimensionalBS,Friedman1977AnAF,PENG1995438}, Locally-weighted regression~\citep{Baird1993ReinforcementLW,locallyweighedatekson,Moore1997EfficientLW}, and Kernel-based regression~\citep{Connell1987LearningTC,Dietterich2001BatchVF,kbrl,Xu2006KernelLT,Bhat2012NonparametricAD,BarretoAndr2016PracticalKR}. Despite their solid theoretical footing, these methods, have limited applicability as they are difficult to scale to high-dimensional state and action spaces. More recently, several methods that propose using recurrent neural networks for learning history-based abstractions have enjoyed considerable success in complex computer games~\citep{rnn-hausknecht,JaderbergMCSLSK17,impala,DBLP:conf/iclr/GruslysDAPBM18,ha} however most of these methods have been designed for partially observable environments where use of history-based methods is often necessary. To the best of our knowledge, the only other work where a history-based RL algorithm is used for controlling a MDP is presented by \citet{openai2019learning}. In this work the authors show that using an LSTM-based agent architecture results in superior performance for the object reorientation using robotic arms. However, the authors do not provide a theoretical analysis of their method. 
    % The algorithmic framework in \Cref{sec:algorithm} can be 
    \subsection{ Bisimulation metrics}
     On the theoretical front, our work is closely related to state aggregation techniques based on bisimulation metrics proposed by~\citet{Givan2003EquivalenceNA,Ferns2004MetricsFF,Ferns2011BisimulationMF}. The bisimulation metric is the  fixed point of an operator on the space of semi-metrics defined over the state space of an MDP with Lipschitz value functions. Apart from state aggregation, bisimulation metrics have been used for feature discovery~\citep{Comanici2011BasisFD,Ruan2015RepresentationDF}, and transfer learning~\citep{Castro2010UsingBF}. However, computational impediments have prevented their broad adoption. Our work can be viewed as an alternative to bisimulation for the analysis of history-based state abstractions and deep RL methods. Our work can also be thought of as extension of the DeepMDP framework~\citep{deepmdp} to history-based policies and direct policy search methods. %{\bf{TODO: Verify this with Aditya and Doina}}.
     \subsection{AIS and Agent state}
     The notion of AIS is closely related to the epistemic state recently proposed by \citet{bvrstate}. An epistemic state is a bounded representation of the history. It is updated recursively as the agent collects more information, and is represented as an environment proxy $\Upsilon$ which is learnt by optimising a target/objective function $\chi$. Since $\Upsilon$ is a random variable, its entropy $\mathbb{H}(\Upsilon)$ is used to represent system's uncertainty about the environment. The framework proposed in this paper can considered as a practical way of constructing the system epistemic state where, the AIS $\Ais_\timestep$ represents both the epistemic state and the environment proxy $\Upsilon$, $\aisloss$ represents  $\chi$, and instead of entropy, the constants $\epsilon$, and $\delta$ represent the systems uncertainty about the environment. The study of the AIS framework in the regret minimisation paradigm could help establish a relationship between the $\epsilon$, $\delta$, and $\mathbb{H}(\Upsilon)$, thereby helping designers develop principled algorithms which synthesise ideas like information directed sampling for direct policy search algorithms.
    \subsection{Analysis of RL algorithms with attention mechanism}
    Recently, there has been considerable interest in developing RL algorithms which use attention mechanism/transformer architectures~\citep{BahdanauCB14,Xu2015ShowAA} for learning feature abstractions~\citep{Zambaldi2019DeepRL,Mott2019TowardsIR,Sorokin2015DeepAR,Oh,RitterFSSBR21,ParisottoSRPGJJ20,chen,LoyndFcSH20,TangNH20,PritzelUSBVHWB17}. Attention mechanism extract task relevant information from historical observations and can be used instead of RNNs for processing sequential data~\citep{vaswani}. As we do not impose a functional from on the history compression function $\aisfunction_\timestep(\cdot)$ in \cref{def:ais}, any attention mechanism can be interpreted as history compression function, and one can construct a valid information state by ensuring that the output of the attention mechanism satisfies (P1) and (P2). That being said, even without optimising $\aisloss$, the approximation bound in \cref{thm:ais-dp} still applies for RL algorithms with attention mechanisms, with the caveat that the constants $\epsilon$, and $\delta$ may be arbitrarily large. A thorough empirical analysis of the effect of different attention mechanisms, and the AIS loss on the on the error constants $\epsilon$, and $\delta$ could help us gain a better understanding of the way in which such design choices could influence the learning process.
    \subsection{AIS for POMDPs}
      The concept of an AIS used in this paper is similar to the idea of AIS for POMDPs ~\citep{ais-2, ais-1}. Moreover, the literature also contains several other methods which have enjoyed empirical success in using history-based policies for controlling POMDPs \citep{Isbell,hutte-1,hutter-2,Schaefer2007ARC,Dreamer,Pla-Net}. In principle, one can use any of these methods for controlling MDPs. However, this does not immediately provide a tight bound for the approximation error. The MDP model has more structure than POMDPs, and our goal in this paper is to use this fact to present a tighter analysis of the approximation error.
\section{Conclusion and future work}\label{sec:conclusion}
    This paper presents the design and analysis of a principled approach for learning history-based policies for controlling MDPs. We believe that our approximation bounds can be helpful for practitioners to study the effect of some of their design choices on the solution quality. On the practical side, the proposed algorithm shows favourable results on high-dimensional control tasks. Note that one can also use the bounds in \Cref{thm:ais-dp} to analyse the approximation error of other history-based methods. However, since some of these algorithms do not satisfy \Cref{def:ais}, the resulting approximation error might be arbitrarily large. Such blow-ups in the approximation error could be because the bound itself is loose or the optimality gap is large. This would depend on the specifics of the methods and remains to be investigated. As such, a sharper analysis of the approximation error by factoring in the specific design choices of other methods is an interesting direction for future research. Another interesting direction would be to conduct a thorough empirical evaluation exploring the design choices of history compression functions.

\bibliography{references}

\appendix
\pagenumbering{arabic}
\section*{Appendix}
\section{Proof for Theorem \ref{thm:ais-dp}}\label{sec:proof:thm:ais-dp}

For readability we will restate the theorem statement
      \begin{theorem}
             For any time $\timestep$, any realisation $\sts_\timestep$ of $\State_\timestep$, $\action_\timestep$ of $\Action_\timestep$, let $\history_\timestep = (\sts_{1:\timestep}, \action_{1:\timestep-1})$, and $\ais_\timestep = \aisfunction_\timestep(\history_\timestep)$.
             The worst case difference between $\valuefunction^{\star}$ and $\valuefunction^{\policy{}}_{\timestep}$ is bounded as:
            \begin{align}
                \Delta
                \le 2 \frac{\varepsilon + \discount\delta \kappa_{\mathfrak{F}}(\hat \valuefunction^{\mu}, \hat{f})}{1 - \discount},
            \end{align}
        where, $\kappa_{\mathfrak{F}}(\hat \valuefunction, \hat f) = \sup_{\feature, \action}\rho_{\mathfrak{F}}(\hat \valuefunction(\hat f(\cdot, \feature, \action)))$. and $\rho_{\mathfrak{F}}(\cdot)$ is the Minkowski functional associated with the IPM $\ipm$ as defined in \eqref{eq:minkowski-functional}.
        \end{theorem}
 \begin{proof}\label{proof:thm:ais-dp}
    For this proof we will use the following convention: For a generic history $\history_\timestep \in \historyspace_\timestep$, we assume that $\history_\timestep  = (\sts_{1:\timestep}, \action_{1:\timestep-1})$, moreover, note that $\ais_\timestep = \aisfunction_\timestep(\history_\timestep)$.

   Now from \eqref{eq:ipm-function-diff}, and \Cref{def:ais} for any $\action_\timestep, \sts_\timestep, \ais_\timestep$:
        \begin{align}
            \nonumber  \max_{\history \in \historyspace_\timestep, \action_{\timestep} \in \actionspace}\bigg\vert \cost (\sts_\timestep, \action_\timestep) - \hat \cost(\ais_\timestep, \action_\timestep)\bigg\vert &\leq \epsilon.\\
            \max_{\history \in \historyspace_\timestep, \action_{\timestep} \in \actionspace}\bigg\vert \sum_{\sts_{\timestep+1} \in \statespace} \bigg(\transition(\sts_{\timestep+1}\vert \sts_\timestep,\action_\timestep)\hat{\valuefunction}^{\mu}(\hat{f}(\sts_{_\timestep+1}, \ais_\timestep, \action_\timestep)) - \hat{\transition}(\sts_{\timestep+1}\vert \ais_\timestep,\action_\timestep)\hat{\valuefunction}^{\mu}(\hat{f}(\sts_{\timestep+1}, \ais_\timestep, \action_\timestep))\bigg)\bigg\vert &\leq \delta \rho_{\mathfrak{F}}(\hat{\valuefunction}(\hat f(\cdot, \feature_\timestep, \action_\timestep))).\label{eq:delta}
        \end{align}
       Now using triangle inequality we get:
        \begin{align}
            \Vert \valuefunction^{\star}(\sts_\timestep) - \valuefunction_{\timestep}^{\pi}(\history_\timestep)\Vert_{\infty}
            &\stackrel{(a)}{\leq} \underbrace{\Vert \valuefunction^{\star}(\sts_\timestep) - \hat \valuefunction^{\mu}(\ais_\timestep) \Vert_{\infty}}_{\text{term 1}} +  \underbrace{\Vert \valuefunction_{\timestep}^{\policy{}}(\history_\timestep) - \hat\valuefunction^{\mu}(\ais_\timestep) \Vert_{\infty}}_{\text{term 2}},
        \end{align}
        where $(a)$ follows from triangle inequality.

        We will now proceed by bounding terms 1 and 2 separately
        
        \paragraph{Bounding term 1:}
        % Using inequality $\max f(x) \leq \max \vert f(x) - g(x)\vert + \max g(x)$ we get:
        \begin{align}
           \Vert \valuefunction^{\star}(\sts_\timestep) - \hat \valuefunction^{\mu}(\feature_\timestep) \Vert_\infty
            &\leq \max_{\history \in \historyspace_\timestep}\bigg \vert\max_{\action_{\timestep} \in \actionspace}\bigg[ Q^{\star}(\sts_\timestep, \action_\timestep) - \hat Q^{\mu}(\feature_\timestep, \action_\timestep)\bigg]\bigg\vert, \label{eq:vq-diff-1}
        \end{align}
        Therefore, for any action $\action_\timestep$ 
        \begin{align*}
             \max_{\history \in \historyspace_\timestep}\bigg \vert\max_{\action_{\timestep} \in \actionspace}\bigg[Q^\star(\sts_\timestep, \action_\timestep) - \hat Q^{\mu}(\feature_\timestep, \action_\timestep)\bigg]\bigg\vert 
             &= \max_{\history \in \historyspace_\timestep}\bigg \vert \max_{\action_{\timestep} \in \actionspace}\bigg[\cost(\sts_{\timestep},\action_\timestep) + \discount\sum_{\sts_{\timestep+1} \in \statespace} \transition(\sts_{\timestep+1} \vert \sts_\timestep,\action_\timestep)\valuefunction^{\star}(\sts_{\timestep+1})\\
             &-  \hat\cost(\ais_\timestep,\action_\timestep) - \discount\sum_{\sts_{\timestep+1} \in \statespace} \hat{\transition}(\sts_{\timestep+1} \vert \ais_\timestep,\action_\timestep)\hat \valuefunction^{\mu}(\hat{f}(\sts_{\timestep+1},\ais_\timestep, \action_\timestep))\bigg]\bigg\vert\\
             &\stackrel{(a)}{\leq} \epsilon + \max_{\history \in \historyspace_\timestep, \action_{\timestep} \in \actionspace} \bigg\vert \discount\sum_{\sts_{\timestep+1} \in \statespace} \transition(\sts_{\timestep+1} \vert \sts_\timestep,\action_\timestep)\valuefunction^{\star}(\sts_{\timestep+1})
             - \discount \sum_{\sts_{\timestep+1} \in \statespace}\transition(\sts_{\timestep+1}\vert \sts_\timestep,\action_\timestep)\hat{\valuefunction}^{\mu}(\hat{f}(\sts_{\timestep+1}, \ais_\timestep, \action_\timestep)) \bigg\vert\\
             &+ \max_{\history \in \historyspace_\timestep, \action_{\timestep} \in \actionspace} \bigg\vert \discount\sum_{\sts_{\timestep+1} \in \statespace}\transition(\sts_{\timestep+1}\vert \sts_\timestep,\action_\timestep)\hat{\valuefunction}^{\mu}(\hat{f}(\sts_{\timestep+1}, \ais_\timestep, \action_\timestep)) - \discount\sum_{\sts_{\timestep+1} \in \statespace} \hat{\transition}(\sts_{\timestep+1} \vert \ais_\timestep,\action_\timestep)\hat \valuefunction^{\mu}(\hat{f}(\sts_{\timestep+1},\ais_\timestep, \action_\timestep))\bigg\vert\\
            &\stackrel{(b)}{\leq} \epsilon + \discount \Vert (\valuefunction^{\star}(\sts_\timestep) -\hat{\valuefunction}^{\mu}(\ais_\timestep))\Vert_\infty  +\discount \delta \rho_{\mathfrak{F}}(\hat{\valuefunction}^{\mu}(\hat f(\cdot, \feature_\timestep, \action_\timestep))),\\
            % &\leq \frac{\epsilon + \delta_{\mathfrak{F}}\rho_{\mathfrak{F}}(\hat{\valuefunction})\kappa_{\mathfrak{F}}(\hat f)}{(1-\discount)}.
        \end{align*}
        where $(a)$ from triangle inequality and $(b)$ is due to \eqref{eq:delta}. Now defining $\kappa_{\mathfrak{F}}(\hat \valuefunction, \hat f) = \sup_{\feature, \action}\rho_{\mathfrak{F}}(\hat \valuefunction(\hat f(\cdot, \feature, \action)))$, and substituting the above result in \eqref{eq:vq-diff-1} we get
        \begin{align}
            \vert \valuefunction^{\star}(\sts_\timestep) - \hat \valuefunction^{\mu}(\feature_\timestep) \vert 
            &\leq \frac{\varepsilon + \discount\delta \kappa_{\mathfrak{F}}(\hat \valuefunction^{\mu}, \hat{f})}{1 - \discount}\label{eq:term1}.
        \end{align}
        \paragraph{Bounding term 2:}
        % Using inequality $\max f(x) \leq \max \vert f(x) - g(x)\vert + \max g(x)$ we get:
        \begin{align}
            \Vert \valuefunction_{\timestep}^{\policy{}}(\history_\timestep) - \hat \valuefunction^{\mu}(\feature_\timestep) \Vert_\infty
            &\leq \max_{\history \in \historyspace_\timestep}\bigg \vert\max_{\action_{\timestep} \in \actionspace}\bigg[ Q_{\timestep}^{\policy{}}(\history_\timestep, \action_\timestep) - \hat Q^{\mu}(\feature_\timestep, \action_\timestep)\bigg]\bigg\vert,  \label{eq:vq-diff-2}
        \end{align}
        Therefore, for any action $\action_\timestep$ 
        \begin{align*}
           \max_{\history \in \historyspace_\timestep}\bigg \vert\max_{\action_{\timestep} \in \actionspace}\bigg[ Q^{\policy{}}(\history_\timestep, \action_\timestep) - \hat Q^{\mu}(\feature_\timestep, \action_\timestep)\bigg]\bigg\vert
            &=\max_{\history \in \historyspace_\timestep}\bigg \vert \max_{\action_{\timestep} \in \actionspace}\bigg[\cost(\sts_{\timestep},\action_\timestep) + \discount\sum_{\sts_{\timestep+1} \in \statespace} \transition(\sts_{\timestep+1} \vert \sts_\timestep,\action_\timestep)\valuefunction_{\timestep+1}^{\policy{}}(\history_{\timestep+1}) \\
            &-  \hat\cost(\ais_\timestep,\action_\timestep) - \discount\sum_{\sts_{\timestep+1} \in \statespace} \hat{\transition}(\sts_{\timestep+1} \vert \ais_\timestep,\action_\timestep)\hat \valuefunction^{\mu}(\hat{f}(\sts_{\timestep+1},\ais_\timestep, \action_\timestep))\bigg]\bigg\vert\\
            &\stackrel{(a)}{\leq} \epsilon + \max_{\history \in \historyspace_\timestep, \action_{\timestep} \in \actionspace} \bigg\vert \discount\sum_{\sts_{\timestep+1} \in \statespace} \transition(\sts_{\timestep+1} \vert \sts_\timestep,\action_\timestep)\valuefunction_{\timestep+1}^{\policy{}}(\history_{\timestep+1})
            - \discount \sum_{\sts_{\timestep+1} \in \statespace}\transition(\sts_{\timestep+1}\vert \sts_\timestep,\action_\timestep)\hat{\valuefunction}^{\mu}(\hat{f}(\sts_{\timestep+1}, \ais_\timestep, \action_\timestep)) \bigg\vert\\
            &+ \max_{\history \in \historyspace_\timestep, \action_{\timestep} \in \actionspace} \bigg\vert \discount\sum_{\sts_{\timestep+1} \in \statespace}\transition(\sts_{\timestep+1}\vert \sts_\timestep,\action_\timestep)\hat{\valuefunction}^{\mu}(\hat{f}(\sts_{\timestep+1}, \ais_\timestep, \action_\timestep))
            - \discount\sum_{\sts_{\timestep+1} \in \statespace} \hat{\transition}(\sts_{\timestep+1} \vert \ais_\timestep,\action_\timestep)\hat \valuefunction^{\mu}(\hat{f}(\sts_{\timestep+1},\ais_\timestep, \action_\timestep))\bigg\vert\\
            &\stackrel{(b)}{\leq} \epsilon + \discount \Vert (\valuefunction^{\policy{}}(\history_\timestep) -\hat{\valuefunction}^{\mu}(\ais_\timestep))\Vert_\infty  +\discount \delta \rho_{\mathfrak{F}}(\hat{\valuefunction}^{\mu}(\hat f(\cdot, \feature_\timestep, \action_\timestep))),\\
            % &\leq \frac{\epsilon + \delta_{\mathfrak{F}}\rho_{\mathfrak{F}}(\hat{\valuefunction})\kappa_{\mathfrak{F}}(\hat f)}{(1-\discount)}.
        \end{align*}
        where $(a)$ is from triangle inequality,  $(b)$ is due to \eqref{eq:delta}, with $\kappa_{\mathfrak{F}}(\hat \valuefunction, \hat f) = \sup_{\feature, \action}\rho_{\mathfrak{F}}(\hat \valuefunction(\hat f(\cdot, \feature, \action)))$, and substituting the above result in \eqref{eq:vq-diff-2} we get
        \begin{align}
            \Vert \valuefunction^{\policy{}}_{\timestep}(\history_\timestep) - \hat \valuefunction^{\mu}(\feature_\timestep) \Vert_\infty
            &\leq \frac{\varepsilon + \discount\delta \kappa_{\mathfrak{F}}(\hat \valuefunction^{\mu}, \hat{f})}{1 - \discount}\label{eq:term2}.
        \end{align}
        
        The final result then follows by adding \eqref{eq:term1} and \eqref{eq:term2}.
       \end{proof}
       
\section{Proof for Corollary \ref{THM:TV-BOUND}}\label{sec:tv-proof}
    \begin{lemma}\label{lem:tv}
        If $\hat\valuefunction$ is the optimal value function of the MDP $\hat{\mdp}$ induced by the process $\{\Feature_{\timestep}\}_{\timestep \geq 0}$, then
        \begin{align}
            \spn(\hat\valuefunction) &\leq \frac{\spn(\hat \cost)}{1 - \discount}.
        \end{align}
    \end{lemma}
    \begin{proof}
        The result follows by observing that the per-step reward $ \hat \cost  (\Feature_\timestep, \Action_\timestep) \in [\min(\hat \cost), \max(\hat \cost) ]$. Therefore $\max(\hat\valuefunction) \leq \max(\hat \cost)$ and  $\min(\hat\valuefunction) \geq \min(\hat \cost)$.
    \end{proof}
    
    \begin{corollary}
             If the function class $\mathfrak{F}$ is $\mathfrak{F}^{\text{TV}}$, then $\Delta$ defined in \eqref{eq:sup-v} is upper bounded as:
             \begin{align}
             \Delta
                \le  \frac{2\epsilon}{1-\discount} +\frac{\discount\delta \spn(\hat\cost)} {(1-\discount)^2},
            \end{align}
        \end{corollary}
        \begin{proof}
            From \Cref{def:tv-dist} we know that for the Total variation distance $\rho_{\mathfrak{F}^{\text{TV}}}(\hat\valuefunction)$ = $\spn(\hat\valuefunction)$ and $\kappa(\hat f) = 1$. The result in the corollary then follows from \Cref{lem:tv}.
        \end{proof}
       
\section{Proof for Corollary \ref{THM:LIP-BOUND}} \label{sec:w-proof}
    \begin{definition}\label{def:lipdist}
        For any Lipschitz function $f:(\aisspace, d_{\Ais}) \to (\real, \vert \cdot \vert)$, and probability measures $\nu_1$, and $\nu_2$ on $(\aisspace, d_{\Ais})$
        \begin{align}
            \bigg\vert \int_{\aisspace} f d\nu_1 - \int_{\aisspace} f d\nu_2 \bigg\vert \leq \Vert f \Vert_{L}. d_{\mathfrak{F}^{\text{W}}}(\nu_1, \nu_2) \leq L_{f}d_{\mathfrak{F}^{\text{W}}}(\nu_1, \nu_2),
        \end{align}
        where $L_f$ is the Lipschitz constant of $f$ and $d_{\mathfrak{F}^{\text{W}}}$ is the Wasserstein distance.
    \end{definition}
    % Note that the process $\{\Feature_{\timestep}\}_{\timestep \geq 0}$ is a 
    \begin{definition}\label{def:lipmdp}
        Let $d$ be a metric on the AIS/Feature space $\aisspace$. The MDP $\hat{\mdp}$ induced by the process $\{\Feature_{\timestep}\}_{\timestep \geq 0}$ is said to be $(L_{\hat\cost}, L_{\hat\transition})$ - Lipschitz if for any $\Feature_1, \Feature_2 \in \aisspace$, the reward $\hat{\cost}$ and transition $\hat{\transition}$ of $\hat{\mdp}$ satisfy the following:
        \begin{align}
            \bigg\vert \cost(\Feature_1, \Action) - \cost(\Feature_2, \Action)\bigg\vert &\leq L_{\hat{\cost}} d(\Feature_1, \Feature_2)\\
            d_{\mathfrak{F}^{\text{W}}}(\hat\transition(\cdot \vert\Feature_1,\Action), \hat\transition(\cdot\vert\Feature_2,\Action) &\leq L_{\hat \transition} d(\Feature_1, \Feature_2),
        \end{align}
    \end{definition}
    where $d_{\mathfrak{F}^{\text{W}}}$ is the Wasserstein or the Kantorovitch-Rubinstein distance.
    % \begin{lemma}\label{lem:w-lip}
    %     If the MDP $\hat{\mdp}$ induced by the $\{\Feature_{\timestep}\}_{\timestep \geq 0}$ is $(L_{\hat \cost}, L_{\hat \transition})$ - Lipschitz and $\hat\valuefunction$ is the optimal value function of $\hat{\mdp}$, then:
    %     \begin{align}
    %         L_{\hat \valuefunction} \leq \frac{(1-\discount)L_{\hat\cost}}{(1-\discount L_{
    %         \hat \transition})}.
    %     \end{align}
    % \end{lemma}
    
    \begin{lemma}\label{lem:lipq}
        Let $\hat \valuefunction: \aisspace \to \real$ be $L_{\hat \valuefunction}$ continuous. Define:
        \begin{align*}
            \hat Q(\ais,\action) &= \hat \cost(\ais, \action) + \discount\sum_{\sts'}\hat\transition(
            \sts'\vert \ais, \action)\hat \valuefunction(\hat f(\sts', \ais, \action).
        \end{align*}
        Then $\hat Q$ is $(L_{\hat \cost} + \discount L_{\hat \valuefunction} L_{\hat f} L_{\hat \transition})$-Lipschitz continuous.
    \end{lemma}
    \begin{proof}
        For any action $\action$
        \begin{align}
          \bigg \vert \hat Q(\ais_1, \action) -\hat Q(\ais_2, \action) \bigg\vert &\stackrel{(a)}{\leq} \bigg\vert \hat\cost(\ais_1, \action) - \hat\cost(\ais_2, \action) \bigg\vert + \discount \bigg\vert \sum_{\sts'}\hat \transition(\sts'\vert \ais_1, \action) \hat \valuefunction(\hat f(\sts',\ais_1, \action)) - \hat \transition(\sts' \vert \ais_2, \action) \hat \valuefunction(\hat f(\sts',\ais_2, \action)) \bigg\vert \\
          &\stackrel{(b)}{\leq} (L_{\hat\cost} + \discount L_{\hat \valuefunction} L_{\hat f} L_{\hat \transition}) d(\ais_1, \ais_2),
        \end{align}
        where $(a)$ due to triangle inequality, and $(b)$ follows form \Cref{def:lipdist}, \Cref{def:lipmdp}, and because $\Vert a \circ b \Vert_{L} \leq \Vert a\Vert_{L} \cdot \Vert b \Vert_{L}$.
    \end{proof}
    
    \begin{lemma}\label{lem:qv}
     Let $\hat Q: \aisspace \times \actionspace \to \real$ be $L_{\hat Q}$- Lipschitz continuous, Define
     \begin{align*}
         \hat \valuefunction(\ais) = \max_{\action_{\timestep} \in \actionspace}\hat Q(\ais, \action).
     \end{align*}
     Then $\hat \valuefunction$ is $L_{\hat Q}$ Lipschitz
    \end{lemma}
    \begin{proof}
    Consider $\ais_1, \ais_2 \in \aisspace$, and let $\action_1$ and $\action_2$ denote the corresponding optimal action. Then, 
        \begin{align}
            \hat \valuefunction(\ais_1) - \hat\valuefunction(\ais_2) &= \hat Q(\ais_1, \action_1) - \hat Q(\ais_2, \action_2)\\
            &\stackrel{(a)}{\leq} \hat Q(\ais_1, \action_2) - \hat Q(\ais_2, \action_2)\\
            &\stackrel{(b)}{\leq}L_{\hat Q}d(\ais_1, \ais_2),
        \end{align}
        By symmetry,
        \begin{align*}
             \hat\valuefunction(\ais_2) - \hat \valuefunction(\ais_1) &\leq L_{\hat Q}d(\ais_1, \ais_2).
        \end{align*}
        Therefore,
        \begin{align*}
            \bigg \vert  \hat \valuefunction(\ais_1) - \hat\valuefunction(\ais_2) \bigg\vert &\leq L_{\hat Q}d(\ais_1, \ais_2).
        \end{align*}
    \end{proof}
    
    \begin{lemma}\label{lem:infv}
        Consider the following dynamic program defined in \eqref{eq:ais-dp}:\footnote{We have added $\timestep$ as a subscript to denote the computation time \ie, the time at which the respective function is updated.}
         \begin{align*}
                \hat Q_\timestep(\ais_\timestep, \action_\timestep) &= \hat \cost(\ais_\timestep, \action_\timestep) + \discount \sum_{\sts_\timestep \in \statespace}
                \hat \transition(\sts_\timestep|\ais_\timestep,\action_\timestep) \hat \valuefunction(\hat{f}(\ais_\timestep,\sts_\timestep,\action_\timestep)), \ \forall \feature \in \featurespace, \action \in \actionspace \\
                \hat \valuefunction_\timestep(\ais_\timestep) &= \max_{\action \in \actionspace} \hat Q_\timestep(\ais_\timestep,\action_\timestep), \ \forall \feature \in \featurespace \label{eq:ais-dp-2}
            \end{align*}
        Then at any time $\timestep$, we have:
        \begin{align*}
            L_{\hat \valuefunction_{\timestep+1}} & = L_{\hat\cost} + \discount L_{\hat \transition}L_{\hat f}L_{\hat \valuefunction_\timestep}.
        \end{align*}
    \end{lemma}
    
    \begin{proof}
        We prove this by induction. At time $\timestep =1$ $\hat Q_{1}(\ais, \action) = \hat \cost(\ais, \action)$, therefore $ L_{\hat Q_{1}} = L_{\hat \cost}$. Then according to \Cref{lem:qv}, $\hat \valuefunction_1$ is Lipschitz with Lipschitz constant $L_{\hat \valuefunction_1} = L_{\hat Q_1} = L_{\hat\cost}$. This forms the basis of induction. Now assume that at time $\timestep$,  $\hat \valuefunction_{\timestep}$ is $L_{\hat \valuefunction_\timestep}$- Lipschitz. By \Cref{lem:lipq} $\hat Q_{\timestep+1}$ is $L_{\hat \cost} + \discount L_{\hat f}, L_{\hat \transition} L_{\hat \valuefunction_{\timestep}}$. Therefore by \Cref{lem:qv}, $\hat \valuefunction_{(\timestep+1)}$ is Lipschitz with constant:
        \begin{align*}
            L_{\hat\valuefunction_{\timestep+1}} &= L_{\hat \cost} + \discount L_{\hat f}L_{\hat \transition}L_{\hat \valuefunction_\timestep}.
        \end{align*}
    \end{proof}
    
    \begin{theorem}\label{thm:thmlipv}
        Given any $(L_{\hat\cost}, L_{\hat\transition})$- Lipschitz MDP, if $\discount L_{
        \hat \transition}L_{\hat f} \leq 1$, then the infinite horizon $\discount$-discounted value function $\hat \valuefunction$ is Lipschitz continuous with Lipschitz constant 
        \begin{align*}
            L_{\hat \valuefunction} &= \frac{L_{\hat \cost}}{1 - \discount L_{\hat f}L_{\hat \transition}}.
        \end{align*}
    \end{theorem}
    \begin{proof}
        Consider the sequence of $L_{\timestep} = L_{\hat \valuefunction_{\timestep}}$ values. For simplicity write $\alpha = \discount L_{\hat \transition}L_{\hat f}$. Then the sequence $\{L_{
        \timestep}\}_{\timestep \geq 1}$ is given by : $L_1 = L_{\hat \cost}$ and for $\timestep \geq 1$,
        \begin{align*}
            L_{\timestep+1} &= L_{\hat \cost} + \alpha L_{\timestep}, \\
            \text{Therefore,} \\
            L_{\timestep} &= L_{\hat \cost} + \alpha L_{\hat \cost}+ \ldots + \alpha_{\timestep+1} = \frac{1 - \alpha^{\timestep}}{1 - \alpha}L_{\hat\cost}.
        \end{align*}
        This sequence converges if $\vert \alpha \vert \leq 1$. Since $\alpha$ is non-negative, this is equivalent to $\alpha\leq 1$, which is true by hypothesis. Hence $L_\timestep$ is a convergent sequence. At convergence, the limit $L_{\hat \valuefunction}$ must satisfy the fixed point of the recursion relationship introduced in \Cref{lem:infv}, hence,
        \begin{align*}
            L_{\hat \valuefunction}&= L_{\hat\cost} + \discount L_{\hat f}L_{\hat \transition}L_{\hat \valuefunction}.
        \end{align*}
        Consequently, the limit is equal to,
        \begin{align*}
            L_{\hat \valuefunction} = \frac{L_{\hat \cost}}{1 - \discount L_{\hat f}L_{\hat \transition}}.
        \end{align*}
    \end{proof}
     \begin{corollary}
             If $\discount L_{
        \hat \transition}L_{\hat f} \leq 1$ and the function class $\mathfrak{F}$ is $\mathfrak{F}^{\text{W}}$, then $\Delta$ as defined in \eqref{eq:sup-v} is upper bounded as:
            \begin{align}
               \Delta
                \le  \frac{2\epsilon}{(1-\discount)} + \frac{2\discount\delta L_{\hat \cost} }{(1- \discount)(1-\discount L_{\hat f}L_{\hat\transition})},
            \end{align}
        \end{corollary}
        \begin{proof}
            The proof follows from the observation that for $d_{\mathfrak{F}^{\text{W}}}$, $\rho_{\mathfrak{F}^{\text{W}}}$ = $L_{\hat \valuefunction}$, and then using the result from \Cref{thm:thmlipv}.
        \end{proof}
        
\section{Algorithmic Details}
    \subsection{Choice of an IPM:}
        \subsubsection{MMD}\label{sec:mmd-details}
            One advantage of choosing $\ipm$ as the MMD distance is that unlike the Wasserstein distance, its computation does not require solving an optimisation problem. Another advantage is that we can leverage some of their properties to further simplify our computation, as follows:
        
            \begin{proposition}[Theorem 22 ~\citep{sejdinovic}] \label{prop-ipm1}
                Let $\mathcal{X} \subseteq \real^{m}$, and $d_{\mathcal{X},p}: \mathcal{X}\times \mathcal{X} \to \real$ be a metric given by $d_{\mathcal{X},p}(x,x') = \Vert x - x'\Vert^{p}_{2}$, for $p \in (0,2]$. Let $k_p :\mathcal{X} \times \mathcal{X} \to \real$ be any kernel given:
                \begin{align}
                    k_p (x,x') &= \frac{1}{2}(d_{\mathcal{X},p}(x,x_0) + d_{\mathcal{X},p}(x',x_0) - d_{\mathcal{X},p}(x,x')),
                \end{align}
                where $x_0 \in \mathcal{X}$ is arbitrary, and let $\mathcal{U}_p$ be a RKHS kernel with kernel $k_p$ and $\mathfrak{F}_p = \{f \in \mathcal{U}_p : \Vert f \Vert_{\mathcal{U}_p} \geq 1  \}$. Then for any distributions $\nu_1$, $\nu_2 \in \Delta{\mathcal{X}}$, the IPM can be expressed as:
              \begin{align}
                    \ipm(\nu_1,\nu_2) &= \bigg(\expecun{}[d_{\mathcal{X},p}(X_1, W_1)] - \frac{1}{2}\expecun{}[d_{\mathcal{X},p}(X_1, X_2)] - \frac{1}{2}\expecun{}[d_{\mathcal{X},p}(W_1, W_2)]\bigg)^{\frac{1}{2}},\label{eq:prop-ipm1}
                \end{align}
                where $X_1,X_2$, and $W_1,W_2$ are i.i.d. samples from $\nu_1$ and $\nu_2$ respectively.
            \end{proposition}
            The main implication of \Cref{prop-ipm1} is that, instead of using \eqref{eq:prop-ipm1}, for $p\in (0,2]$ we can use the following as a surrogate for $d_{\mathfrak{F}_p}$:
                \begin{align}
                     \int_\mathcal{X}\int_\mathcal{X}\Vert x_1 - w_1 \Vert_{2}^{p}\nu_1(dx_1)\nu_2(dw_1) - \frac{1}{2}\int_\mathcal{X}\int_\mathcal{X}\Vert w_1 - w_2 \Vert_{2}^{p}\nu_{2}(dw_1)\nu_{2}(dw_1). \label{eq:ipm-surrogate}
                \end{align}
                Moreover, according to \citet{Sriperumbudur} for n identically and independently distributed (i.i.d) samples $\{X_i\}_{i=0}^{n} \sim \nu_1$ an unbiased estimator of \eqref{eq:ipm-surrogate} is given as:
                \begin{align}
                    \frac{1}{n}\sum_{i=1}^{n}\int_\mathcal{X}\Vert X_i -w_1\Vert_{2}^{p}\nu_1 d(w_1)  - \frac{1}{2}\int_\mathcal{X}\int_\mathcal{X}\Vert w_1 - w_2 \Vert_{2}^{p}\nu_1(dw_1)\nu_2(dw_2). \label{eq:ipm-surrogate-2}
                \end{align}
            
            We implement a simplified version of the surrogate loss in \eqref{eq:ipm-surrogate-2} as follows:
            \begin{proposition}[~\citep{ais-1}]\label{prop-imp2}
            %  \begin{proposition}\label{prop-imp2}
                Given the setup in \cref{prop-ipm1} and $p=2$, Let $\nu_2(\aisparams)$ be a parametric distribution with mean $m$ and let $X \sim \nu_1$, then the gradient $\grad_\aisparams (m_\aisparams - 2X)^{\top}m_\aisparams$ is an unbiased estimator of $\grad_\aisparams d_{\mathfrak{F}_{2}}(\alpha, \nu_\aisparams)^{2}$
            \end{proposition}
            \begin{proof}
                Let $X_1,X_2 \sim \nu_1$, and $W_1,W_2 \sim \nu_2(\aisparams)$
                \begin{align}
                    \therefore \grad_{\aisparams} d_{\mathfrak{F}_2}(\nu_1,\nu_2(\aisparams))^2 &= \grad_{\aisparams} \bigg[\expecun{}\Vert X_1 - W_1 \Vert_{2}^{2} - \frac{1}{2}\expecun{}\Vert X_1 - X_2 \Vert_{2}^{2} - \frac{1}{2}\expecun{}\Vert W_1 - W_2 \Vert_{2}^{2}\bigg ]\\
                    &\stackrel{(a)}{=}\grad_\aisparams \bigg[ \expecun{}\Vert W_1\Vert_{2}^{2} - 2\expecun{}\Vert X_1\Vert^{\top}\expecun{}\Vert W_1 \Vert \bigg]\label{eq:mmd-grad}, 
                \end{align}
                where $(a)$ follows from the fact that $X$ does not depend on $\aisparams$, which simplifies the implementation of the MMD distance. 
            \end{proof}
          In this way we can simplify the computation of $d_\mathfrak{F}$ using a parametric stochastic kernel approximator and MMD metric.
           
          Note that when are trying to approximate a continuous distribution we can readily use the loss function \eqref{eq:mmd-grad} as long as the mean $m_\aisparams$ of $\nu_2(\aisparams)$ is given in closed form. The AIS loss is then given as:
          \begin{align}
                 \aisloss(\aisparams) &= \frac{1}{\Timestep}\sum_{t = 0}^{\Timestep}\bigg( \lambda (f_{\hat{\cost}}(\Ais_{\timestep}, \Action_\timestep; \aisparams)  - \cost(\State_\timestep, \Action_\timestep))^{2} + (1-\lambda)(m^{\State_\timestep}_{\aisparams} - 2\State_\timestep)^{\top}m^{\State_\timestep}_{\aisparams}  \bigg),\label{eq:mmd-ais-loss-2}
            \end{align}
            where $m^{\State_\timestep}_{\aisparams}$ is obtained using the from the transition approximator, ~\ie, the mapping $f_{\hat\transition}(\aisparams): \aisspace \times \actionspace \to \real$.
        
        \subsubsection{Wasserstein Distance}\label{sec:wass-details}
        The the KL-divergence between two densities $\nu_1$ and $\nu_2$ on for any $X \in \mathcal{X} \subset \real^{m} $ is defined as:
        \begin{align}
            d_{\text{KL}}(\nu_1 \Vert \nu_2) &= \int_{\mathcal X} \log(\nu_1(x))\nu_1(dx) - \int_{\mathcal X} \log(\nu_2(x))\nu_1(dx)
        \end{align}
        Moreover, if $\mathcal{X}$ is bounded space with diameter $D$, then the relation between the Wasserstein distance $d_{\mathfrak{F}^{\text{W}}}$, Total variation distance $d_{\mathfrak{F}^{\text{TV}}}$, and the KL divergence is given as :
        \begin{align}
            d_{\mathfrak{F}^{\text{W}}}(\nu_1, \nu_2)\leq D d_{\mathfrak{F}^{\text{TV}}}(\nu_1, \nu_2)\stackrel{(a)}{\leq} \sqrt{2d_{\text{KL}}(\nu_1\Vert \nu_2)},
        \end{align}
        where, $(a)$ follows from the Pinsker's inequality. Note that in \eqref{eq:pgt-loss} we use $d_{\mathfrak{F}}^{2}$. Therefore, we can use a (simplified) KL-divergence based surrogate objective given as:
        \begin{align}
            \int_{\mathcal{X}}\log(\nu_2(x;\aisparams)) \nu_1(dx),
        \end{align}
        where we have dropped the terms which do not depend on $\aisparams$. Note that the above expression is same as the cross entropy between $\nu_1$ and $\nu_2$ which can be effectively computed using samples.
        In particular, if we get $\Timestep$ i.i.d samples from $\nu_1$, then, 
        \begin{align}
            \frac{1}{\Timestep}\sum_{i=0}^{\Timestep}\log(\nu_2(x_i;\aisparams))\label{eq:klwass}
        \end{align}
        is an unbiased estimator of $\int_{\mathcal X}\log(\nu_2(x;\aisparams)) \nu_1(dx)$.
        
        The KL divergence based AIS loss is then given as:
          \begin{align}
                \aisloss(\aisparams) &= \frac{1}{\Timestep}\sum_{t = 0}^{\Timestep}\bigg( \lambda (f_{\hat{\cost}}(\Ais_{\timestep}, \Action_\timestep; \aisparams)  - \cost(\State_\timestep, \Action_\timestep))^{2} + (1-\lambda)\log(\hat \transition(\State_\timestep;\aisparams))  \bigg),\label{eq:w-ais-loss-2}
            \end{align}

\section{Experimental Details}\label{sec:experiment-details}
    \begin{table}[!htbp]
        \begin{center}
            \begin{tabular}{|c|l|c|}
                \hline
                \multirow{9}{*}{Common}&Optimiser& Adam \\
                & Discount Factor $\discount$ & 0.99 \\
                % & GAE parameter $\lambda$& 0.95\\
                &Inital standard deviation for the policy &0.0\\
                & PPO-Epochs & 12\\
                & Clipping Coefficient & 0.2\\
                & Entropy-Regulariser& 0\\
                & Batch Size & 512\\
                & Episode Length & 2048\\
                \hline
                \multirow{3}{*}{AIS generator}& History Compressor& GRU \\
                & Hidden layer dimension & 256 \\
                & Step size & 1.5e-3\\
                & $\lambda$ & 0.3\\
                \hline
                \multirow{3}{*}{Actor}
                & Step size & 3.5e-4\\
                & No of hidden layers& 1 \\
                & Hidden layer Dimension& 32\\
                % \hline
                % \multirow{3}{*}{Critic} 
                % & Step size & 1.5e-3\\
                % & No of hidden layers& 1 \\
                % & Hidden layer Dimension& 32\\
                \hline
            \end{tabular}
        \end{center}
        \caption{Hyperparameters}\label{tab:hyperparams}
    \end{table}

    \subsection{Environments}
         Our algorithms are evaluated on MuJoCo~\citep[mujoco-py version 2.0.2.9 ]{Todorov2012MuJoCoAP} via OpenAI gym~\citep[version 0.17.1]{gym} interface, using the v2 environments. The environment, state-space, action space, and reward function are not modified or pre-processed in any way for easy reproducibility and fair comparison with previous results. Each environment runs for a maximum of 2048 time steps or until some termination condition and has a multi-dimensional action space with values in the range of (-1, 1), except for Humanoid which uses the range of (-0.4, 0.4).

    \subsection{Hyper-parameters}
        \Cref{tab:hyperparams} contains all the hyper-parameters used in our experiments.
        Both the policy and AIS networks are trained with Adam optimiser~\citep{adam}, with a batch size of 512. We follow~\citet{Raichuk2021WhatMF}'s recommended protocol for training on-policy policy based methods, and perform 12 PPO updates after every policy evaluation subroutine. To ensure separation of time-scales the step size of the AIS generator and the policy network is set to $1.5\text{e}^{-3}$ and $3.5\text{e}^{-4}$ respectively. Hyper-parameters of our approach are searched over a grid of values, but an exhaustive grid search is not carried out due to prohibitive computational cost. We start with the recommended hyper-parameters for the baseline implementations and tune them further around promising values by an iterative process of performing experiments and observing results. 
        
        For the state-based RNN baseline we have tuned the learning rate over a grid of values starting from 1e-4 to 4e-4 and settled on 3.5e-4 as it achieved the best performance. Similarly the hidden layer size set to 256 as it is observed to achieve best performance. For the feed-forward baselines we use the implementation by OpenAI baselines~\citep{baselines} with their default hyper-parameters.

\end{document}